\documentclass[runningheads]{llncs}

\usepackage{eccv}

\usepackage{eccvabbrv}

\usepackage{graphicx}
\usepackage{booktabs}

\usepackage[accsupp]{axessibility}  

\usepackage{graphicx}
\usepackage{booktabs}

\usepackage{pifont}%

\newcommand{\rom}[1]{\uppercase\expandafter{\romannumeral #1\relax}}

\usepackage{xcolor}

\usepackage{calc}

\usepackage[accsupp]{axessibility}  %
\usepackage{colortbl}
\usepackage{tabulary}
\usepackage{tabularx}
\usepackage{etoolbox}
\usepackage{multirow}
\usepackage{cuted}
\usepackage[percentage]{overpic}
\usepackage{booktabs}
\usepackage{pgfplots}
\pgfplotsset{compat=newest}%
\usepackage{algorithm}
\usepackage{algpseudocode}
\usepackage{amsmath}
\usepackage{bbding}

\usepackage{stackrel}

\usepackage{tikz}
\usetikzlibrary{calc}
\usetikzlibrary{decorations.pathmorphing}
\usetikzlibrary{patterns,arrows,automata,positioning,decorations,shapes,calc}

\tikzstyle{normalvertex}=[circle,fill=white,draw=black]
\tikzstyle{emptyvertex}=[draw,circle,minimum size=7pt,inner sep=0pt]
\tikzstyle{tinyvertex}=[draw,circle,minimum size=3pt,inner sep=0pt]
\tikzstyle{thickedge}=[draw,gray!60,line width=1.6pt,-]

\usepackage{pgfplots}
\usepgfplotslibrary{colorbrewer}

\pgfdeclarelayer{background}
\pgfsetlayers{background,main}

\usetikzlibrary{shapes}
\usetikzlibrary{arrows.meta,backgrounds,automata, chains}
\usetikzlibrary{positioning,shapes,matrix,calc}
\tikzstyle{vertex}=[circle, draw, fill=gray!80!white,thick,scale=1.2]
\tikzstyle{edge}=[draw=black, thick,-]

\usepackage{bm}

\definecolor{purple}{RGB}{147,7,204}
\definecolor{blue}{RGB}{10,153,201}
\definecolor{orange}{RGB}{254,128,41}
\definecolor{gray}{RGB}{239,240,241}
\definecolor{pink}{RGB}{254,15,127}
\definecolor{green}{RGB}{140,211,89}

\definecolor{color1}{RGB}{254,15,127}
\definecolor{color2}{RGB}{10,153,201}
\definecolor{color3}{RGB}{194,145,162}
\definecolor{color4}{RGB}{254,128,41}
\definecolor{color5}{RGB}{254,191,185}
\definecolor{color6}{RGB}{110,231,169}
\definecolor{color7}{RGB}{245,221,66}

\newcommand{\X}{\mathcal{X}}
\newcommand{\Y}{\mathcal{Y}}
\newcommand{\XY}{\mathcal{XY}}
\newcommand{\YX}{\mathcal{YX}}
\newcommand{\T}{\mathrm{T}}
\newcommand{\tr}{\mathrm{tr}}
\newcommand{\F}{\mathrm{F}}

\usepackage[pagebackref,breaklinks,colorlinks,citecolor=eccvblue]{hyperref}

\usepackage{orcidlink}

\begin{document}


\title{Coarse-to-Fine: A Hybrid Self-Supervised Method for Non-rigid 3D Shape Matching} 



\titlerunning{Coarse-to-Fine Hybrid Self-Supervised Shape Matching}

\author{
Feifan Luo\inst{1,2}\orcidlink{0009-0002-0574-970X} \and
Ting Li\inst{3}\orcidlink{0009-0004-8591-1213} \and
Zhao Li\inst{2}\orcidlink{0000-0002-5056-0351} \and
Hongyang Chen\inst{2}\orcidlink{0000-0002-7626-0162}\thanks{Corresponding author.}
}

\authorrunning{Feifan Luo et al.}

\institute{Zhejiang University, China \and
Zhejiang Lab, China \and Wenzhou University of Technology, China \\
\email{{\tt\small luoff@zju.edu.cn}\quad {\tt\small dr.h.chen@ieee.org}}
}



\maketitle

\begin{abstract} 
Non-rigid 3D shape matching is a fundamental task in computer vision and graphics. In this paper, we propose a hybrid self-supervised method based on a coarse-to-fine strategy, which ensures consistency between the coarse mapping and the refined correspondence produced by our refinement module. The architecture features a dual-branch design, consisting of two symmetric functional map learning streams: one based on the Laplacian basis and the other utilizing the elastic basis. Extensive experiments show that our approach not only maintains computational efficiency, but also achieves state-of-the-art performance across a variety of challenging scenarios, including non-isometric deformations and topological noise. Finally, we rigorously demonstrate that contrastive energies promote feature discrimination. Furthermore, integrating these energies with existing methods yields consistent improvements, validating the overall efficacy of our approach. Our code is available at \url{https://github.com/LuoFeifan77/Coarse-to-Fine-Hybrid-Self-Supervised-Matching}.

\end{abstract}
 
\section{Introduction}
\label{sec:intro}
Establishing accurate correspondences between non-rigid 3D shapes is a long-standing challenge in computer vision and graphics, which has a wide range of applications, including shape analysis~\cite{loncaric1998survey}, shape generation~\cite{egger20203d}, deformation transfer~\cite{sumner2004deformation}, and shape interpolation~\cite{eisenberger2021neuromorph}.

Deep functional maps~\cite{litany2017deep} have emerged as a preferred solution for non-rigid shape matching, paving the way for significant advancements in near-isometric shape matching~\cite{donati2020deep, halimi2019unsupervised, roufosse2019unsupervised, Ayguen2020}, non-isometric shape matching~\cite{eisenberger2020deep, eisenberger2021neuromorph, HU2023101189, donati2022deep,li2022learning}, multi-shape matching~\cite{cao2022unsupervised, Sun_2023_ICCV, eisenberger2023g}, and partial shape matching~\cite{attaiki2021dpfm, ehm2024partial, ehm2024geometrically}. 

However, these methods~\cite{roufosse2019unsupervised, Ayguen2020, donati2020deep, HU2023101189, donati2022deep, li2022learning} primarily rely on geometric prior constraints (e.g., orthogonality) to supervise pointwise or functional maps, while overlooking self-supervised prior constraints, leading to suboptimal performance. Although incorporating the coupling loss~\cite{Cao2023} to promote functional map properness~\cite{ren2021discrete} enhances performance, it inevitably incurs substantial computational overhead. This is because these methods either necessitate solving linear optimization problems to estimate functional maps~\cite{Cao2023, bastian2024hybrid, attaiki2023understanding, Sun_2023_ICCV, luo2025deep, donati2022deep} or rely on fine-tuning techniques~\cite{Cao2023, bastian2024hybrid, cao2024spectral, le2024integrating} during inference to bolster results.

To address these limitations, we propose a novel and efficient hybrid self-supervised method for non-rigid shape matching. We first introduce \textit{a coarse-to-fine framework} for functional map learning. The core idea is to align the coarse map with the refined map, and it consists of two main components: (1) the proposal of three different self-supervised prior constraints, which facilitate the alignment of the coarse map and the refined map from spatial, spectral, and spatial-spectral perspectives; (2) the introduction of a general refinement method, which can be applied not only to generate both refined pointwise and functional maps but also to non-orthogonal settings. Based on the aforementioned framework, \textit{a hybrid self-supervised method} is composed of two fully symmetric branches: the intrinsic branch, which employs a functional map learning module based on the orthonormal Laplacian basis, and the extrinsic branch, which uses a functional map learning module based on the non-orthogonal elastic basis, while eliminating the time-consuming least-squares solvers and fine-tuning techniques. Extensive experiments demonstrate that our method outperforms existing state-of-the-art methods in both matching accuracy and runtime. Finally, we rigorously prove that contrastive energies promote feature discrimination and lead to consistent improvements when integrated with other methods, thereby validating the effectiveness of our proposed framework.

The main contributions are summarized as follows:
\begin{itemize}
    \item We propose a general coarse-to-fine strategy, which establishes a unified bridge between functional map learning frameworks and map refinement methods.
    \item We present a robust and efficient hybrid self-supervised method for non-rigid 3D shape matching.
    \item Extensive experiments demonstrate that our approach achieves excellent performance across challenging benchmarks, while maintaining remarkable computational efficiency.
    \item We provide a rigorous proof that contrastive energies induce feature discrimination. Moreover, our proposed energies deliver consistent performance gains when integrated into baseline methods.
\end{itemize}

\section{Related Work}
\label{sec:rw}
Non-rigid 3D shape matching is a basic problem that has been extensively studied. For a comprehensive overview of the field, readers are referred to surveys such as \cite{sahilliouglu2020recent,deng2022survey,liu2024spectral,zhuravlev2026non}.

\subsection{Axiomatic Functional Maps}
Numerous axiomatic approaches~\cite{besl1992method,Ovsjanikov2012,vestner2017kernelmatching,Xiang_2020_CVPR,aflalo2016spectral,bronstein2006generalized} have been proposed to address the shape matching problem. Among these, the functional map framework~\cite{Ovsjanikov2012} has gained significant prominence due to its inherent flexibility and efficiency, which has inspired extensive subsequent works~\cite{ren2018continuous,Ren2019,Melzi2019,
maggioli2024rematching, Huang2020, Hu2021, Gao_2021_CVPR, fan2022coherent, 2022SmoothNonRigidShapeMatchingviaEffectiveDirichletEnergyOptimization, Donati2022, 2023ElasticBasis,vigano2025nam,Gautam2021}. However, axiomatic functional map methods rely heavily on the quality of handcrafted features~\cite{sun2009concise,Aubry2011The,salti2014shot,liu2024awedd,li2021anisotropic}, resulting in unsatisfactory performance in challenging settings. 

\subsection{Deep Functional Maps}
FMNet~\cite{litany2017deep}, also known as deep functional maps, effectively overcomes the bottlenecks of axiomatic functional map approaches~\cite{Ovsjanikov2012} by learning local features via neural networks. Subsequent approaches~\cite{halimi2019unsupervised, Ayguen2020,donati2020deep} followed this principle, constraining pointwise or functional maps to enhance feature representation~\cite{roufosse2019unsupervised, sharma2020weakly}, where the geometric prior constraints like bijectivity and orthogonality~\cite{Ren2019} have been widely adopted. Recently, the remarkable success of functional maps in ensuring properness~\cite{Cao2023,Sun_2023_ICCV} has brought dual-branch functional map learning architectures~\cite{cao2024revisiting,cao2024spectral,bastian2024hybrid,luo2025deep,le2024integrating,attaiki2023understanding} and a single-branch self-supervised network~\cite{magnet2024memory}. Moreover, approaches based on hybrid bases~\cite{bastian2024hybrid,li2026mdnd}, diffusion models~\cite{pierson2025diffumatch,zhuravlev2025denoising}, neural adjoint  maps~\cite{vigano2025nam},
unsupervised contrastive learning~\cite{luo2026unsupervised}, 
basis learning~\cite{luo2026feature}, and other studies~\cite{weber2024finsler,bracha2024unsupervised,bracha2024wormhole,yona2025neural,roetzer2024spidermatch,xie2025echomatch} have gradually been applied to the field of shape matching. 

Despite advancements, existing methods either demonstrate poor matching performance~\cite{roufosse2019unsupervised, sharma2020weakly, Ayguen2020, donati2020deep, HU2023101189, donati2022deep, li2022learning} in challenging scenarios, such as non-isometric deformations and cross-dataset generalization, or rely on time-consuming linear system solvers~\cite{donati2020deep, Cao2023, donati2022deep, li2022learning, cao2024spectral, bastian2024hybrid, cao2022unsupervised, cao2024revisiting, cao2023self, Sun_2023_ICCV, cao2024synchronous, attaiki2023understanding}, or fine-tuning techniques~\cite{Cao2023, cao2024spectral, bastian2024hybrid, cao2024revisiting, cao2024synchronous}. In contrast, we propose an effective and efficient hybrid self-supervised method for non-rigid shape matching, which outperforms these methods in both matching accuracy and computational efficiency. Furthermore, we demonstrate that the self-supervised losses introduced in previous works~\cite{magnet2024memory, cao2024revisiting} are special cases of the contrastive energy proposed in our study. Finally, the contrastive energies introduced in our method consistently improve the performance of other functional map learning techniques, such as ULRSSM~\cite{Cao2023}.

\section{Background} 
We begin by providing a brief overview of the basic pipeline for deep functional maps, directing interested readers to relevant literature~\cite{Ovsjanikov2012, Cao2023, bastian2024hybrid,zhuravlev2026non} for further details. The notation used throughout the paper is summarized in Table~\ref{tab: symbol desc}.

\begin{table}[h!t]
\centering
\caption{Summary of the symbol description.}
\scalebox{0.80}{
\begin{tabular}{lcl}
\toprule
Symbol     &         & Description           \\ \midrule
$\X, \Y$  &   & 3D shapes (triangle mesh) with $|V_\mathcal{X}|$, $|V_\Y|$ vertices      \\ 
$\Lambda_{\X}$  &  & $\mathbb{R}^{k \times k}$ eigenvalue matrix of shape $\X$    \\
$\Phi_{\X}$ &  & $\mathbb{R}^{|V_\mathcal{X}| \times k}$ eigenfunctions  matrix   \\
$\Phi^{\dagger}_{\X}$ &  & $\mathbb{R}^{k \times |V_\mathcal{X}|}$ Moore-Penrose inverse of $\Phi_{\X}$     \\

$A_{\X}$ &  & $\mathbb{R}^{|V_\mathcal{X}| \times |V_\mathcal{X}|}$ mass matrix.   \\
$A_{\X,k}$ &  & $\mathbb{R}^{k \times k}$ reduced mass matrix.   \\

$F_{\X}$ &  & $\mathbb{R}^{|V_\mathcal{X}| \times d}$ vertex-wise features of shape ${\X}$, $d$ denotes the feature dimension     \\
$\Pi_{\YX}$ &  & $\mathbb{R}^{|V_\mathcal{Y}| \times |V_\mathcal{X}| }$ point-wise map between shapes $\Y$ and $\X$    \\
$C_{\XY}$ &  & $\mathbb{R}^{k \times k}$ (spectral) functional maps matrix obtained by solving a linear system Eq.\eqref{equ: desc and reg}    \\
$\hat{C}_{\XY}$ &  & $\mathbb{R}^{k \times k}$ spatial functional maps matrix obtained from $\Pi_{\YX}$   \\
$||\cdot||_{\mathrm{F}}$ &  & Frobenius norm for Laplacian eigensystem \\
$||\cdot||_{\mathrm{HS}}$ &  & Hilbert–Schmidt norm for elastic eigensystem.    \\
$\{g_s(\cdot)\}^{S}_{s=1}$ &  & a set of filter functions \\
\bottomrule
\end{tabular}}
\label{tab: symbol desc}
\end{table}

\subsection{Functional Map Learning Pipeline} 
 
Given a pair of non-rigid 3D shapes denoted as ${\mathcal{X}}$ and ${\mathcal{Y}}$, the common functional map learning frameworks mainly consist of five main stages:

(1) \textbf{Precomputation.} Compute the first $k$ eigenfunctions $\Phi_{\mathcal{X}}, \Phi_{\mathcal{Y}}  $, eigenvalues $\Lambda_{\mathcal{X}}, \Lambda_{\mathcal{Y}} $ in matrix notation via generalized eigendecomposition, respectively. For example, the orthonormal Laplacian eigensystem $\{\Phi^{LB}_{\X}, \Lambda^{LB}_{\X}\}$ and the non-orthogonal elastic eigensystem $\{\Phi^{EL}_{\X}, \Lambda^{EL}_{\X}\}$ can be obtained by decomposing the Laplacian operator~\cite{pinkall1993computing} and elastic thin-shell energy~\cite{wirth2011continuum}.

(2) \textbf{Feature extraction.} The vertex features $F_{\mathcal{X}}$ and $F_{\mathcal{Y}}$ are extracted by a feature extractor network $\mathcal{F}_\Theta$, respectively, with shared parameters $\Theta$. Among numerous feature extraction methods, DiffusionNet~\cite{sharp2022diffusionnet} remains the predominant choice for the majority of functional map learning approaches.

(3) \textbf{Functional map computation.} Functional maps have two computation methods. The first is the \textit{definition-based} approach~\cite{Ovsjanikov2012}, which is obtained by solving a linear optimization problem, specifically:
\begin{small}
\begin{equation}\label{equ: desc and reg}
   {C}_{\mathcal{XY}} = \mathop{\arg\min}\limits_{{C}_{\mathcal{XY}}} \left\|   {C}_{\mathcal{XY}} \Phi_{\mathcal{X}}^{\dagger}  F_{\mathcal{X}} - \Phi_{\mathcal{Y}}^{\dagger}  F_{\mathcal{Y}} \right\| + \mu E_{reg}({C}_{\mathcal{XY}}), 
\end{equation}
\end{small}
where $E_{reg} = \left\| C_{\mathcal{XY}}\Lambda_{\mathcal{X}} - \Lambda_{\mathcal{Y}} C_{\mathcal{XY}} \right\|$ denotes the regularization term,  $\mu$ is a hyperparameter. Here, 
the choice of norm is determined by the type of basis functions $\Phi$ employed. For the orthonormal Laplacian eigensystem, Eq.~\eqref{equ: desc and reg} is computed using the Frobenius norm $||\cdot||_{\mathrm{F}}$, whereas for the non-orthogonal elastic eigensystem, the Hilbert–Schmidt norm $||\cdot||_{\mathrm{HS}}$ is adopted~\cite{bastian2024hybrid}.

The second is the \textit{projection-based} approach~\cite{Ovsjanikov2012}, where the pointwise map is projected onto the spectral domain, expressed as:
\begin{equation}\label{equ: compute C by Pi}
{\hat{C}}_\mathcal{XY}=\Phi_{\mathcal{Y}}^\dagger\Pi_\mathcal{YX}\Phi_{\mathcal{X}},
\end{equation}
where the pointwise map satisfies $F_{\Y} = \Pi_{\YX} F_{\X}$. 

To avoid ambiguity, the functional map $C_{\XY}$ derived from Eq.\eqref{equ: desc and reg} is termed the (spectral) functional map, while that explicitly computed via pointwise map, i.e., Eq.~\eqref{equ: compute C by Pi}, is designated as the \textit{spatial} functional map~\cite{Sun_2023_ICCV,xu2025sedfmnet} (or the proper functional map in some references~\cite{ren2021discrete,magnet2024memory,pierson2025diffumatch}). 

(4) \textbf{Loss functions.} During the training phase, axiomatic prior constraints, such as bijectivity and orthogonality~\cite{Ren2019}, are typically applied as structural penalties, namely,
\begin{equation}\label{equ: fmap loss}
    L_{bi}  =\left\| {C}_\mathcal{XY} {C}_\mathcal{YX} -{I}\right\|, L_{or}  = \left\|  {C}_\mathcal{XY} {C}_\mathcal{XY}^{\mathrm{T}} -{I}\right\|,   
\end{equation}
or the coupling~\cite{Cao2023} between spatial and spectral functional maps, i.e., 
\begin{equation}\label{equ: couple loss}
    L_{co}  =\left\| {C}_\mathcal{XY} - {\hat{C}}_\mathcal{XY}\right\|.   
\end{equation}
Additionally, the aforementioned loss terms can be extended to bidirectional constraints by incorporating reverse mapping terms. 

However, most existing functional map learning methods~\cite{Ayguen2020,zhuravlev2025denoising,donati2022deep,li2022learning} mostly rely on geometric prior constraints while ignoring self-supervised prior constraints, leading to subpar performance in challenging scenarios, such as non-isometric deformations and matching with topological noise. On the other hand, they rely on time-consuming computational modules, such as the linear system solver~\cite{donati2020deep,attaiki2023understanding} in Eq.~\eqref{equ: desc and reg} and fine-tuning techniques~\cite{Cao2023,cao2024spectral,bastian2024hybrid}, which significantly increase the computational cost. In contrast, we propose a novel, efficient, and effective hybrid self-supervised method that achieves outstanding performance in both matching accuracy and computational cost. 


\begin{figure*}[!ht]
	\centering
	\includegraphics[width=0.7\linewidth]{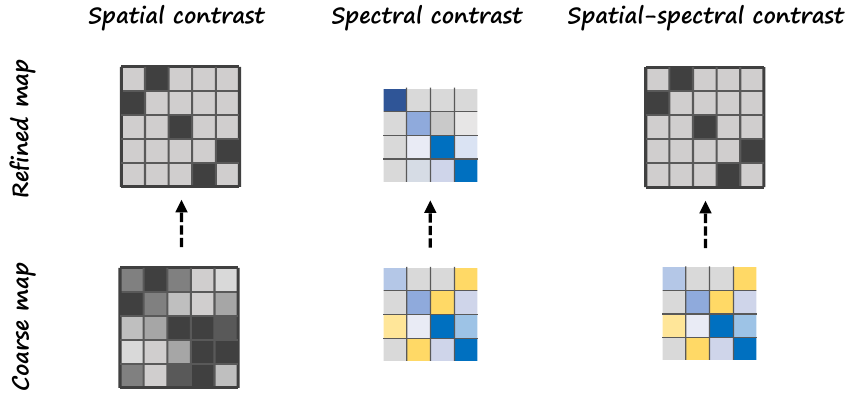}
    \caption{Comparison of the three coarse-to-fine contrastive energy terms. From left to right: spatial contrastive energy, spectral contrastive energy, and spatial-spectral contrastive energy.
    }\label{fig:contrastive_energies_main_text}
\end{figure*}

\section{Coarse-to-Fine Guided Functional Maps}\label{sec:coarse-to-fine}
This section introduces the coarse-to-fine strategy tailored for the functional map framework, which aims to align the coarse map more closely with the refined map. We begin by defining three types of self-supervised prior constraint terms: spatial, spectral, and spatial-spectral contrastive energies, each of which promotes the alignment of the coarse and refined maps from different directions in Section~\ref{sec:contrastive_energies}. We then present a general refinement method that generates refined functional and pointwise maps, applicable to both orthogonal and non-orthogonal basis functions in Section~\ref{sec:g_map_refinement}.

\subsection{Contrastive Energy}\label{sec:contrastive_energies}
The essence of the coarse-to-fine strategy lies in enforcing the coarse maps to converge toward the refined maps, where the latter serves as a high-fidelity guidance. To this end, we formulate a series of contrastive energy constraints based on this coarse-to-fine hierarchy, see Fig.~\ref{fig:contrastive_energies_main_text}.

\textbf{Spatial contrast.} Given a coarse pointwise map $\Pi_\mathcal{YX}$, the pointwise map contrastive energy is defined as:
\begin{equation}\label{equ: pointcontrastive_cross}
    \| \Pi_\mathcal{YX} - \Pi^{ref}_\mathcal{YX}\|,   
\end{equation}
where $\Pi^{ref}_{\mathcal{YX}}$ represents the refined pointwise map. Since the refined pointwise maps $\Pi^{ref}_{\mathcal{YX}}$ encode higher-quality correspondences, they can be regarded as a proxy for ground truth, guiding the optimization of the coarse pointwise maps $\Pi_{\mathcal{YX}}$.  


If the pointwise map represents a self-mapping relationship within the shape itself, denoted as $\Pi_\mathcal{YY}$, then the refined map $\Pi^{ref}_{\mathcal{YY}}$ reduces to the identity mapping ${I_{\Y}} \in \mathbb{R}^{|V_{\Y}| \times |V_{\Y}|}$ due to the self-evident correspondence of vertex features, leading to
\begin{equation}\label{equ: pointcontrastive_self}
    \| \Pi_\mathcal{YY} - {I_{\Y}}\|.  
\end{equation}



Interestingly, the self-supervised constraints introduced in earlier works~\cite{cao2024revisiting,magnet2024memory} are special cases of our spatial contrastive energies, with additional details provided in the supplementary materials.

\textbf{Spectral contrast.}
Parallel to the spatial contrastive regularization, we incorporate a coarse-to-fine representation for the spectral functional map $C_\mathcal{XY}$, denoted as
\begin{equation}\label{equ: functionalcontrastive_cross}
    \| C_\mathcal{XY} - C^{ref}_\mathcal{XY}\|.   
\end{equation}

Unlike the self spatial contrastive energy Eq.~\eqref{equ: pointcontrastive_self}, since spectral functional maps are computed from a linear system Eq.~\eqref{equ: desc and reg} via the least-square method, the resulting $ C_\mathcal{YY}$ is necessarily equal to $I$. Consequently, we omit this trivial contrastive constraint.

\textbf{Spatial-spectral contrast.}
The coupling constraint Eq.~\eqref{equ: couple loss}, which promotes properness~\cite{Cao2023,Sun_2023_ICCV}, has proven highly effective in challenging shape matching tasks. However, it specifically aligns the coarse functional maps with the coarse pointwise maps, thereby ensuring what we term as \textit{coarse properness}. In contrast, we propose a novel spatial-spectral contrastive energy to promote \textit{fine properness}, namely, 
\begin{equation}\label{equ: couple loss_pointfine}
    \left\| {C}_\mathcal{XY} - \Phi_\mathcal{Y}^\dagger\Pi^{ref}_\mathcal{YX}\Phi_\mathcal{X}\right\|. 
\end{equation}
The above equation can be interpreted as aligning spectral functional maps with refined pointwise maps.

On the other hand, we also leverage the refined spectral  functional map ${C}_\mathcal{XY}^{ref}$ to facilitate the optimization of the coarse pointwise maps, i.e., 
\begin{equation}\label{equ: couple loss_functionalfine}
    \left\| \Pi_\mathcal{YX} - \Phi_\mathcal{Y}{C}^{ref}_\mathcal{XY}\Phi^\dagger_\mathcal{X} \right\|. 
\end{equation}

Our coarse-to-fine strategy establishes a unified bridge between deep functional map frameworks and map refinement methods, by employing the aforementioned contrastive energies as self-supervised losses to enhance map quality. On the other hand, this integration allows various established axiomatic refinement techniques~\cite{vestner2017kernelmatching,ren2018continuous,Melzi2019,Hu2021} to generate refined maps.

\subsection{Map Refinement}\label{sec:g_map_refinement}

Despite the proposal of numerous axiomatic map refinement techniques~\cite{Melzi2019,2023ElasticBasis}, they lack a unified framework capable of refining both pointwise and functional maps simultaneously. 
Inspired by~\cite{luo2025deep}, we introduce a more generalized refinement paradigm that is equally effective for both pointwise and functional map representations, even under non-orthogonal basis functions.  

\textbf{Functional map refinement.} We extend the refinement method proposed in~\cite{luo2025deep}—specifically, the general multi-scale Laplacian commutativity term—to the context of non-orthogonal basis functions. For additional background and detailed information, we strongly recommend that readers refer to the original paper.
The general multi-scale Laplacian commutativity term is as follows:
\begin{equation}
E({C}_{\mathcal{XY}}) = \sum^{S}_{s=1}  \| {C}_{\mathcal{XY}}g_{s}(\Lambda_\mathcal{X}) - g_{s}(\Lambda_\mathcal{Y}){C}_{\mathcal{XY}}\|^2_{\mathrm{HS}},
\end{equation}
where $ \{g_{s}\}^{S}_{s=1}$ denote the filter functions. 

Although one could compute this following the approach for solving standard Laplacian commutativity in the previous study~\cite{bastian2024hybrid}, such a computation is complex and incurs significant computational overhead. Instead, we recommend adopting a strategy similar to the prior work~\cite{luo2025deep}. Specifically, we assume that one of the functional maps is already obtained (denoted as ${C}^{init}_{\mathcal{XY}}$), then, 
\begin{equation}\label{equ:computer C with fixed C_init}
E({C}_{\mathcal{XY}}) = \sum^{S}_{s=1}  \| {C}_{\mathcal{XY}}g_{s}(\Lambda_\mathcal{X}) - g_{s}(\Lambda_\mathcal{Y}){C}^{init}_{\mathcal{XY}}\|^2_{\mathrm{HS}}.
\end{equation}
Consequently, we only need to solve for the other functional map.



\newtheorem{lem1}[theorem]{Lemma}
\begin{lem1}\label{pro: remark_1}
The regularization term $E({C}_{\mathcal{XY}})$ can be formulated as:
\begin{equation*}
{C}^{ref}_{\mathcal{XY}} = \arg \min_{{C}_{\mathcal{XY}}} \sum^{S}_{s=1} \| \sqrt{A_{\Y, k}}  \left ({C}_{\mathcal{XY}}g_s(\Lambda_\mathcal{X}) - g_s(\Lambda_\mathcal{Y}){C}^{init}_{\mathcal{XY}} \right )\sqrt{A^{-1}_{\X, k}}\|^{2}_{\mathrm{F}},
\end{equation*}
where $A_{\X, k} = \Phi_{\X}^{\mathrm{T}}A_{\X}\Phi_{\X}$.

Moreover, if the set of filter functions $\{g_s(\lambda)\}_{s=1}^S$ satisfy Parseval condition that 
	$\sum_{s} g_s^{2}\left(\lambda\right) \equiv 1, \forall \lambda,$
then the functional map in Eq.~\eqref{equ:computer C with fixed C_init} can be obtained via
\begin{equation}\label{eq: solve spectral functional map}
\begin{aligned}
    {C}^{ref}_\mathcal{XY} = \sum_{s}  g_s\left(\Lambda_{\mathcal{Y}}\right) {C}^{init}_\mathcal{XY}  g_s\left(\Lambda_{\mathcal{X}}\right). 
 \end{aligned}
\end{equation}
\end{lem1}

\begin{proof}
The first statement expresses the HS-norm in terms of the Frobenius norm. For the second claim, the formulation of Eq.~\eqref{eq: solve spectral functional map} is analogous to that in reference~\cite{luo2025deep}. Detailed proofs are provided in supplementary materials.
\end{proof}

Clearly, Eq.~\eqref{eq: solve spectral functional map} can be interpreted as refining the coarse functional map ${C}^{init}_\mathcal{XY}$ using multiple sets of filters, analogous to signal denoising. 

\textbf{Pointwise map refinement.} Similarly, we employ the same strategy to generate refined pointwise maps. First, we compute the spatial functional map $\hat{C}_{\XY} = \Phi_\mathcal{Y}^\dagger\Pi_\mathcal{YX}\Phi_\mathcal{X} $ and then refine it via Eq.~\eqref{eq: solve spectral functional map}, yielding the refined spatial functional map $\hat{C}^{ref}_{\XY}$. 
Then, we utilize a common strategy originating from ZoomOut~\cite{Melzi2019}, which converts the refined spatial functional map $ \hat{C}^{ref}_{\mathcal{XY}}$ into a refined pointwise map by minimizing the following  energy:
\begin{equation}\label{equ:compute refined pointwise map}
{\Pi}^{ref}_{\mathcal{YX}} = \arg \min_{{\Pi}_{\mathcal{YX}}} \left\|\Phi_{\mathcal{Y}} - {\Pi}_{\mathcal{YX}}\Phi_{\mathcal{X}} (\hat{C}^{ref}_\mathcal{XY})^{*}\right\|_{\mathrm{HS}}, 
\end{equation}
where $(\hat{C}^{ref}_\mathcal{XY})^{*} = A^{-1}_{\X,k} (\hat{C}^{ref}_\mathcal{XY})^{\mathrm{T}} A_{\Y,k}$ denotes the \textit{adjoint operator} of $\hat{C}^{ref}_\mathcal{XY}$. 
The above equation can be solved via nearest neighbor search~\cite{2023ElasticBasis}, i.e., 

\begin{equation}\label{equ: solve elastic pointwise map}
\Pi^{ref}_{\mathcal{YX}} = NN \left ( ({A^{1/2}_{\Y,k}} \Phi^{\dagger}_\mathcal{Y}A^{-1}_{\Y})^{\mathrm{T}},( {A^{1/2}_{\Y,k}} \hat{C}^{ref}_{\mathcal{XY}} \Phi^{\dagger}_\mathcal{\X}A^{-1}_{\X})^{\mathrm{T}} \right).
\end{equation}

In the orthonormal setting, since $A_{\X,k} = I$ and $A_{\Y,k} = I$, the adjoint operator of $\hat{C}^{ref}_\mathcal{XY}$ reduces to its transpose, i.e., $(\hat{C}^{ref}_\mathcal{XY})^{*}$ = $(\hat{C}^{ref}_\mathcal{XY})^{\mathrm{T}}$. Additionally, with $\Phi^{\dagger}_\mathcal{Y}A^{-1}_{\Y} = \Phi_{\Y}^{\mathrm{T}}$, Eq.~\eqref{equ: solve elastic pointwise map} is simplified to 
\begin{equation}\label{equ: solve Laplacian pointwise map}
\Pi^{ref}_{\mathcal{YX}} = NN(\Phi_\mathcal{Y},\Phi_\mathcal{X}(\hat{C}^{ref}_{\mathcal{XY}})^\mathrm{T}).
\end{equation}

\begin{algorithm}[h!t]
	\caption{Map refinement}
	\label{power1}
    \small
	\begin{algorithmic} 
		\State \textbf{Input}:  $C_\mathcal{XY}$, $\Pi_\mathcal{YX}$, $\{g_s(\Lambda_{\mathcal{X}})\}_{s=1}^S$, $\{g_s(\Lambda_{\mathcal{Y}})\}_{s=1}^S$
		\State \textbf{Output}:  $C^{ref}_\mathcal{XY}$ and $\Pi^{ref}_\mathcal{YX}$ 
        \State $L_2$-normalization ensures $\sum_{s} g_s^{2}\left(\Lambda_{\mathcal{X}}\right) \equiv 1$ and $\sum_{s} g_s^{2}\left(\Lambda_{\mathcal{Y}}\right) \equiv 1$.
		\State Estimate refined functional map:  ${C}^{ref}_\mathcal{XY} = \sum_{s}  g_s\left(\Lambda_{\mathcal{Y}}\right) {C}_\mathcal{XY}  g_s\left(\Lambda_{\mathcal{X}}\right)$.  
        \State Estimate refined pointwise map: (1): $\hat{C}^{ref}_\mathcal{XY} = \sum_{s}  g_s\left(\Lambda_{\mathcal{Y}}\right)\hat{C}_\mathcal{XY}  g_s\left(\Lambda_{\mathcal{X}}\right)$ with $\hat{C}_{\mathcal{XY}}=\Phi_\mathcal{Y}^\dagger\Pi_\mathcal{YX}\Phi_\mathcal{X}$. (2) Recover $\Pi^{ref}_\mathcal{YX}$ from Eq.~\eqref{equ:compute refined pointwise map}.
	\end{algorithmic}\label{alg:map refinement}
\end{algorithm}

\textbf{Filter normalization.} If the set of filters does not satisfy the Parseval framework~\cite{hammond2011wavelets} (i.e., $\sum_{s} g_s^{2}\left(\lambda\right) \neq 1, \exists \lambda$), the refinement process necessitates matrix inversion~\cite{luo2025deep}. To circumvent numerical instability arising from matrix inversion, we impose an $L_2$ norm constraint on the filters $\{g_s(\lambda)\}_{s=1}^S$ to ensure they satisfy the Parseval condition. This approach not only facilitates computation but also enhances numerical robustness.

We summarize the generalized refinement strategy for both functional and pointwise maps in Algorithm~\ref{alg:map refinement}.

\section{Hybrid Self-Supervised Learning Shape Matching}\label{sec:hybrid method}

\begin{figure*}[h!t]
	\centering
	\includegraphics[width=0.9\linewidth]{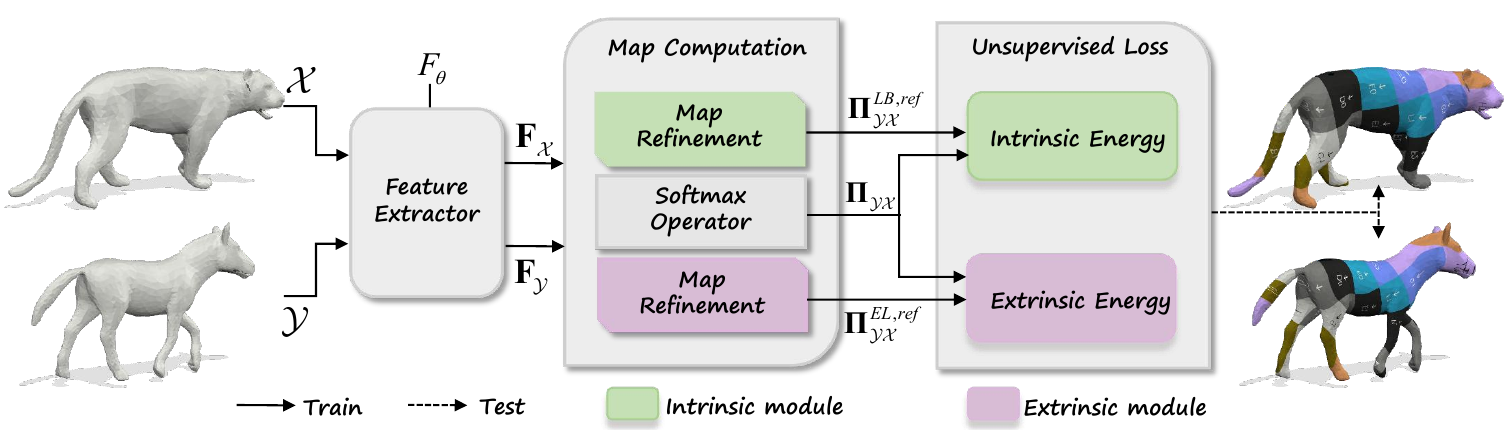}
	\caption{An overview of our method. (1) Feature extraction: Utilize DiffusionNet~\cite{sharp2022diffusionnet} to extract vertex representations ${F}_{\X}$ and ${F}_{\Y}$ from $\mathcal{X}$ and $\mathcal{Y}$, respectively.
    (2) Map computation: Employ the softmax operator and the refinement Algorithm~\ref{alg:map refinement} to generate the coarse soft map $\Pi_{\YX}$ and the refined hard maps $\Pi^{LB,ref}_{\YX}$ and $\Pi^{EL,ref}_{\YX}$, respectively.
    (3) Unsupervised loss: Apply intrinsic and extrinsic energies (Eq.~\eqref{equ: lb total loss} and Eq.~\eqref{equ: el total loss}) to supervise the network learning. 
    (4) Map recovery. Use Eq.~\eqref{equ: nnsearch for refine} and Eq.~\eqref{equ: hybrid nnsearch with map refine} to recover the pointwise maps during inference.
    }\label{fig:our_network}
\end{figure*}

In this section, we propose a novel and efficient hybrid self-supervised learning method for non-rigid 3D shape matching, a dual-branch framework that bypasses the computationally expensive least-squares solver.
This framework consists of an \textit{intrinsic} branch, which constructs functional maps using the Laplacian basis in Section~\ref{sec:intrinsic_branch}, and an \textit{extrinsic} branch, which utilizes an Elastic basis for functional map learning in Section~\ref{sec:extrinsic_branch}. Each branch incorporates two distinct computational pipelines: a refinement stream and unsupervised losses. The network architecture is illustrated in Fig.~\ref{fig:our_network}.

\subsection{Eliminating Least-Squares Optimization}
Solving the functional map optimization problem via least-squares, as shown in Eq.~\eqref{equ: desc and reg}, has become a standard component in many state-of-the-art methods\cite{Cao2023,cao2024revisiting,cao2024synchronous,cao2022unsupervised,luo2025deep,donati2022deep,li2022learning,cao2024spectral,bastian2024hybrid,le2024integrating}. However, the least-squares solver introduces significant computational overhead. For instance, HybridFMaps~\cite{bastian2024hybrid} requires solving a dense $k^2 \times k^2$ linear system (where $k$ is the number of eigenvectors, typically between 100 and 200); detailed runtimes are provided in supplementary materials. Furthermore, least-squares solvers can suffer from numerical instability and require higher-dimensional feature inputs~\cite{magnet2024memory}. Consequently, although we have designed a series of regularization terms Eq.~\eqref{equ: functionalcontrastive_cross}, Eq.~\eqref{equ: couple loss_pointfine}, and Eq.~\eqref{equ: couple loss_functionalfine} for spectral functional maps that could potentially enhance performance, we deliberately bypass these potential gains in favor of computational efficiency. Instead, we establish a highly efficient shape matching framework based \textit{solely} on pointwise map constraints.

\subsection{Coarse Map} 
The coarse map refers to the initial mapping generated prior to any refinement operations. To ensure both differentiability and computational efficiency, after extracting shape features ${F}_{\mathcal{X}} $ and ${F}_{\mathcal{Y}}$, we follow several state-of-the-art methods~\cite{eisenberger2021neuromorph,Cao2023,luo2025deep} by employing a softmax operator to compute the differentiable pointwise map matrix, defined as:
\begin{equation}\label{eq: compute coarse map}
    \Pi_{\mathcal{YX}} = \mathrm{Softmax}({F}_{\mathcal{Y}} {F}^{\mathrm{T}}_{\mathcal{X}}/\tau),
\end{equation}
where the element at position ($j$, $i$) represents the probability of correspondence between the $j$-th point on $\mathcal{Y}$ and the $i$-th point on $\mathcal{X}$, and $\tau$ is the temperature factor to determine the softness of the matrix.

\subsection{Intrinsic Branch}\label{sec:intrinsic_branch}
The Laplacian basis encodes essential intrinsic spectral information and has been widely adopted in functional map-based methods. Motivated by its efficacy, we utilize the spectral  basis to construct the intrinsic self-supervised learning branch of our framework. 

\textbf{Non-differentiable refined map.}
The refined map is designed to provide optimization guidance for the coarse soft map, therefore it does not strictly require differentiability. On the other hand, an ideal soft correspondence matrix $\Pi_{\mathcal{YX}}$ should approximate a binary correspondence matrix. This implies achieving optimal discriminativity, where the mapping is concentrated on the most similar candidate points while maintaining maximum separation from the remaining low-similarity or dissimilar points. 

We first compute the initial binary pointwise map matrix based on a nearest neighbor search, i.e., 
\begin{equation}\label{equ: nnsearch for refine}
    \Pi^{init}_{\mathcal{YX}} = NN(F_\mathcal{Y}, F_\mathcal{X}),
\end{equation}
where $\Pi^{init}_{\mathcal{YX}} \in \{0,1\}^{|V_\mathcal{Y}| \times |V_\mathcal{X}|}$. Subsequently, we apply the Algorithm~\ref{alg:map refinement} to derive the refined binary  pointwise map $\Pi^{LB,ref}_{\mathcal{YX}}$.

\textbf{Unsupervised loss.} Unsupervised loss functions are essential for learning functional maps. Our unsupervised loss design consists of two components: the widely used geometric prior constraints~\cite{Ren2019,ren2021discrete} and the self-supervised constraints proposed previously.

Geometric prior constraints impose structural penalties on the mapping, such as  bijectivity and orthogonality. However, these constraints fail to penalize the image of $\Pi_{\mathcal{YX}}$ that lies outside the span of $\Phi_{\mathcal{Y}}$. Inspired by~\cite{ren2021discrete}, we employ variants of orthogonality and bijectivity penalties to constrain the coarse map ${\Pi}_{\mathcal{YX}}$, formulated as:

\begin{equation}\label{equ: formulated orthogonality loss}
L^{LB}_{or}  = \left\|\Phi^{LB}_{\mathcal{Y}} - {\Pi}_{\mathcal{YX}}\Phi^{LB}_{\mathcal{X}} (\hat{C}^{LB}_\mathcal{XY})^{\mathrm{T}}\right\|_{\mathrm{F}}   
\end{equation} 
and
\begin{equation}\label{equ: formulated bijectivity loss}
    L^{LB}_{bi}  = \left\|\Phi^{LB}_{\mathcal{Y}} - {\Pi}_{\mathcal{YX}}\Phi^{LB}_{\mathcal{X}} \hat{C}^{LB}_\mathcal{YX}\right\|_{\mathrm{F}}, 
\end{equation}
respectively, where $\Phi^{LB}_{\X} \in \mathbb{R}^{ |V_{\X}| \times k^{LB}}$.

However, directly imposing these constraints on coarse maps would lead to a significant numerical discrepancy between different losses, as the orthogonality Eq.~\eqref{equ: formulated orthogonality loss}  and bijectivity terms Eq.~\eqref{equ: formulated bijectivity loss} are defined in $\mathbb{R}^{N_\Y \times k}$ while the spatial contrastive losses Eq.~\eqref{equ: pointcontrastive_cross} and Eq.~\eqref{equ: pointcontrastive_self} reside in $\mathbb{R}^{N_\Y \times N_\X}$ and $\mathbb{R}^{N_\Y \times N_\Y}$, respectively. This imbalance could cause network training to fail. To address this, we multiply by the corresponding Laplacian basis, which not only ensures numerical consistency across loss terms but also has no adverse impact on the optimization of the coarse map, namely, 
\begin{equation}\label{equ: couple loss_pointmap_cross}
    L^{LB}_{cross} = \left\| \Pi_\mathcal{YX}\Phi^{LB}_\mathcal{X} - \Pi^{LB,ref}_\mathcal{YX}\Phi^{LB}_\mathcal{X}\right\|_\mathrm{F}. 
\end{equation}
On the other hand, for self-spatial contrast energy, we also leverage the fine pointwise map $I_\mathcal{Y}$ to facilitate the optimization of the coarse maps, i.e., 
\begin{equation}\label{equ: couple loss_pointmap_self}
    L^{LB}_{self} = \left\| \Pi_\mathcal{YY}\Phi^{LB}_\mathcal{Y} - \Phi^{LB}_\mathcal{Y}\right\|_\mathrm{F}. 
\end{equation}
The unsupervised losses of the intrinsic branch are expressed as:
\begin{equation}\label{equ: lb total loss}
    L^{LB}_{total}  = \theta^{LB}_{cross} L^{LB}_{cross} + \theta^{LB}_{self} L^{LB}_{self} + \theta^{LB}_{bi} L^{LB}_{bi} + \theta^{LB}_{or} L^{LB}_{or},
\end{equation}
where $\theta^{LB}$ denotes the corresponding weight.

\subsection{Extrinsic Branch}\label{sec:extrinsic_branch}
As the functional map learning pipeline on the extrinsic branch mirrors the intrinsic branch in its underlying formulation, we omit repetitive details for the sake of brevity and focus on the primary distinctions.

\textbf{Non-differentiable refined map.} Similarly, 
by employing Algorithm~\ref{alg:map refinement} with the same initial pointwise map, we can obtain the refined binary pointwise maps, denoted as $\Pi^{EL,ref}_{\mathcal{YX}}$.

\textbf{Unsupervised loss.} To impose a structured penalty on elastic maps, we introduce, for the first time, novel orthogonality and bijectivity loss terms, denoted as:
\begin{equation}\label{equ: elastic orthogonality loss}
     L^{EL}_{or}  = \left\|\Phi^{EL}_{\mathcal{Y}} - {\Pi}_{\mathcal{YX}}\Phi^{EL}_{\mathcal{X}} (\hat{C}^{EL}_\mathcal{XY})^{*}\right\|_{\mathrm{HS}} 
\end{equation}
and
\begin{equation}\label{equ: elastic bijectivity loss}
     L^{EL}_{bi}  = \left\|\Phi^{EL}_{\mathcal{Y}} - {\Pi}_{\mathcal{YX}}\Phi^{EL}_{\mathcal{X}} \hat{C}^{EL}_\mathcal{YX}\right\|_{\mathrm{HS}},  
\end{equation}
where $\Phi^{EL}_{\X} \in \mathbb{R}^{ |V_{\X}| \times k^{EL}}$.

These are reformulated in terms of the standard Frobenius norm, denoted as:
\begin{equation}\label{equ: formulated elastic orthogonality loss}
     L^{EL}_{or}  = \left\| A^{1/2}_{\Y} \left ( \Phi^{EL}_{\mathcal{Y}} - {\Pi}_{\mathcal{YX}}\Phi^{EL}_{\mathcal{X}} (\hat{C}^{EL}_\mathcal{XY})^{*}  \right ) A^{-1/2}_{\Y,k}\right\|_{\mathrm{F}} 
\end{equation}
and
\begin{equation}\label{equ: formulated elastic bijectivity loss}
     L^{EL}_{bi}  = \left\|  A^{1/2}_{\Y} \left (  \Phi^{EL}_{\mathcal{Y}} - {\Pi}_{\mathcal{YX}}\Phi^{EL}_{\mathcal{X}} (\hat{C}^{EL}_\mathcal{YX}) \right ) A^{-1/2}_{\Y,k}\right\|_{\mathrm{F}}, 
\end{equation}
respectively. 

Analogously, for the spatial contrastive regularizer defined on elastic bases, we have
\begin{equation}\label{equ: elasic cross spatial loss}
    L^{EL}_{cross} = \left\| \Pi_\mathcal{YX}\Phi^{EL}_\mathcal{X} -\Pi^{EL,ref}_\mathcal{YX}\Phi^{EL}_\mathcal{X}\right\|_\mathrm{HS} 
\end{equation}
 and
\begin{equation}\label{equ: elasic self spatial loss}
    L^{EL}_{self} = \left\| \Pi^{EL}_\mathcal{YY}\Phi^{EL}_\mathcal{Y} - \Phi^{EL}_\mathcal{Y}\right\|_\mathrm{HS}. 
\end{equation}
Expressing them by the Frobenius norm, namely, 
\begin{equation}\label{equ: formulated elasic cross spatial loss}
    L^{EL}_{cross} = \left\| A^{1/2}_{\Y} \left ( \Pi_\mathcal{YX}\Phi^{EL}_\mathcal{X} -\Pi^{EL,ref}_\mathcal{YX}\Phi^{EL}_\mathcal{X} \right ) A^{-1/2}_{\X,k}\right\|_\mathrm{F} 
\end{equation}
 and
\begin{equation}\label{equ: formulated elasic self spatial loss}
    L^{EL}_{self} = \left\| A^{1/2}_{\Y} \left ( \Pi^{EL}_\mathcal{YY}\Phi^{EL}_\mathcal{Y} - \Phi^{EL}_\mathcal{Y} \right ) A^{-1/2}_{\Y,k}\right\|_\mathrm{F}, 
\end{equation}
respectively. 
Then, the unsupervised losses of the extrinsic branch are represented as:
\begin{equation}\label{equ: el total loss}
    L^{EL}_{total}  = \theta^{EL}_{cross} L^{EL}_{cross} + \theta^{EL}_{self} L^{EL}_{self} + \theta^{EL}_{bi} L^{EL}_{bi} + \theta^{EL}_{or} L^{EL}_{or}. 
\end{equation}

Finally, overall losses $L_{total} = L^{LB}_{total} + L^{EL}_{total} $ and the detailed proofs of Eq.~\eqref{equ: formulated elastic orthogonality loss},
~\eqref{equ: formulated elastic bijectivity loss},~\eqref{equ: formulated elasic cross spatial loss}, and~\eqref{equ: formulated elasic self spatial loss} are provided in supplementary materials.

\subsection{Inference}
Inspired by~\cite{Cao2023}, employing distinct recovery strategies during the inference stage yields more robust results, a practice widely adopted in existing state-of-the-art approaches~\cite{le2024integrating,bastian2024hybrid,cao2024synchronous,cao2024revisiting}. For non-isometric matching scenarios, we directly recover pointwise maps from the learned feature embedding space using the nearest neighbor algorithm, namely, $\Pi_{\mathcal{YX}} = NN(F_\mathcal{Y}, F_\mathcal{X})$. 
For near-isometric matching scenarios, we recover pointwise maps from the hybrid basis space by applying the proposed map refinement to the obtained $\Pi_{\mathcal{YX}}$, i.e., 
\begin{small}
\begin{equation}\label{equ: hybrid nnsearch with map refine}
    \Pi^{ref}_{\mathcal{YX}} = NN \left ( \Phi^{LB}_\mathcal{Y} \Vert ({A^{1/2}_{\Y,k}} \Phi^{EL\dagger}_\mathcal{Y}A^{-1}_{\Y})^{\mathrm{T}}, \Phi^{LB}_\mathcal{X}(\hat{C}^{LB,ref}_{\mathcal{XY}})^\mathrm{T})\Vert( {A^{1/2}_{\Y,k}} \hat{C}^{EL,ref}_{\mathcal{XY}} \Phi^{EL,\dagger}_\mathcal{\X}A^{-1}_{\X})^{\mathrm{T}} \right), 
\end{equation}
\end{small}
where $\Vert$ denotes matrix concatenation.

\section{Experimental Results}

\subsection{Datasets}\label{sec: dataset} 
We extensively evaluate our method on widely used challenging benchmarks, including
\begin{itemize}
    \item \textit{Near-isometry.} The near-isometric datasets comprise remeshed FAUST(F for short)~\cite{ren2018continuous}, SCAPE~\cite{ren2018continuous}, and 
    challenging SHREC'19~\cite{melzi2019shrec}. Moreover, anisotropic meshing datasets FAUST~\cite{Cao2023}(F\_a for short) and SCAPE~\cite{Cao2023} were used to evaluate robustness across different discretizations. 
    \item \textit{Non-isometry.} Commonly employed non-isometric datasets include SMAL~\cite{smalzuffi20173d} and DT4D-H~\cite{2022SmoothNonRigidShapeMatchingviaEffectiveDirichletEnergyOptimization}.     
    DT4D-H offers two configurations: intra-class and inter-class. We omit the intra-class setting because it is overly simple, thereby precluding further evaluation of the method's performance. Consequently, we retain SMAL and the most challenging DT4D-H inter-class setting. 
    \item \textit{Topological noise.} The TOPKIDS dataset~\cite{lahner2016shrec} is employed to evaluate our method’s robustness to topological noise.
\end{itemize}
The training and testing protocols on SMAL follow the same procedure as previous related studies~\cite{donati2022deep,zhuravlev2025denoising}, and the remaining datasets use the same splits as those in many earlier works~\cite{li2022learning, Cao2023}.


\subsection{Baselines}
We extensively compare our method with existing non-rigid deformable shape matching methods, which we categorize as follows:
\begin{itemize}
    \item \textit{Axiomatic approaches}, including ZoomOut \cite{Melzi2019}, Smooth Shells \cite{eisenberger2020smooth}, DiscreteOp \cite{ren2021discrete}, and MWP \cite{Hu2021}.
    \item \textit{Supervised approaches}, FMNet~\cite{litany2017deep}, GeomFMaps~\cite{donati2020deep}.
    \item \textit{Unsupervised approaches}, 
    including Deep Shells~\cite{eisenberger2020deep},  
    DUO-FMNet~\cite{donati2022deep},  AttentiveFMaps~\cite{li2022learning},
    RFMNet~\cite{HU2023101189},
    ULRSSM~\cite{Cao2023}, DiffZO~\cite{magnet2024memory}, HybridFMaps~\cite{bastian2024hybrid}, DeepFAFM~\cite{luo2025deep}, and DenoiseFMaps~\cite{zhuravlev2025denoising}. 
\end{itemize}
 
\begin{table}[h!t]
\centering
\caption{ Evaluating the matching results across various benchmarks, including near-isometric shape matching, cross-dataset generalization (FAUST, SCAPE, and SHREC'19),  anisotropic meshing (F\_a and S\_a), non-isometric shape matching (SMAL and DT4D-H), and matching with topological noise (TOPKIDS), respectively. The numbers in the table are mean geodesic errors ($\times 100$). \textbf{Bold}: Best. \underline{Underline}: Runner-up.}
\scalebox{0.8}{
\begin{tabular}{lccccccccccccc}
    \toprule
Train               & \multicolumn{3}{c}{FAUST} & \multicolumn{3}{c}{SCAPE} & \multicolumn{2}{c}{FAUST} & \multicolumn{2}{c}{SCAPE} & \multicolumn{1}{c}{\multirow{2}{*}{SMAL}}  & \multicolumn{1}{c}{\multirow{2}{*}{DT4D-H inter}} & \multicolumn{1}{c}{\multirow{2}{*}{TOPKIDS}}
\\ \cmidrule(lr){2-4} \cmidrule(lr){5-7} \cmidrule(lr){8-9} \cmidrule(lr){10-11} 
Test                & F        & S   & S19    & F        & S   & S19    & F\_a      & S\_a  & F\_a    & S\_a  & \multicolumn{1}{c}{}                      & \multicolumn{1}{c}{}

\\ \midrule  
                    & \multicolumn{11}{c}{Axiomatic Methods}          \\
ZoomOut~\cite{Melzi2019}             & 6.1         & 7.5   & -     & 6.1         & 7.5 & -  & 8.7         & 15.0        & 8.7         & 15.0  & 47.7                                               & 29.0   & 33.7   \\
SmoothShells~\cite{eisenberger2020smooth}        & 2.5         & 4.7   & -     & 2.5         & 4.7  & -    & 5.4         & 5.0         & 5.4         & 5.0    & 34.9                                        & 6.3   & 35.5 \\
DiscreteOp~\cite{ren2021discrete}          & 5.6         & 13.1   & -     & 5.6        & 13.1 & -  & 6.2         & 14.6       & 6.2       & 14.6   & 36.1                                           & 27.6     & 10.8     \\
MWP~\cite{Hu2021}                 & 3.1         & 4.1    & -     & 3.1          & 4.1  & -     & 8.2         & 8.7       & 8.2         & 8.7    & 20.9                                       &   25.4    &  5.7  \\ \hline
                    & \multicolumn{11}{c}{Supervised Methods}   \\                    
FMNet~\cite{litany2017deep}               & 11.1         & 30.0    & -     & 33.0          & 17.0  & -   & 42.0         & 43.0     & 43.0          & 41.0    & -                
                    & 38.0   &- \\             
GeomFMaps~\cite{donati2020deep}           & 2.6         & 3.4    & 9.9     & 3.0          & 3.0  & 12.2   & 3.2         & 3.8    & 8.4     & 3.1     & 4.3                                
                   & 4.1    &-   \\
\hline
                    & \multicolumn{11}{c}{Unsupervised Methods}                                                                     \\
Deep Shells~\cite{eisenberger2020deep}         & 1.7         & 5.4  & 27.4     & 2.7         & 2.5    & 23.4   & 12.0        & 16.0       & 15.0         & 10.0    & 21.4                                          & 31.1   & 13.7  \\
DUO-FMNet~\cite{donati2022deep}           & 2.5         & 4.2   & 6.4     & 2.7        & 2.6  & 8.4    & 3.0         & 4.4        & 3.1         & 2.7     & 4.8                                    & 15.8   & - \\
AttentiveFMaps~\cite{li2022learning}      & 1.9         & 2.6  & 6.4      & 2.2        & 2.2     & 9.9  & 2.4         & \underline{2.8}        & {2.5}         & 2.3   & 4.4                                         & 11.6     & 23.4  \\
RFMNet~\cite{HU2023101189}               & 1.7         & \textbf{2.3}        & 6.3         & \underline{1.7}     & 2.1      & 6.9  & 3.6         & 2.6        & 3.6         & 3.9       & 4.4                    & 13.9   &-  \\
ULRSSM~\cite{Cao2023}         & {1.6}         & 6.7   & 14.5     &  4.8       & {1.9}   & 18.5   & {2.5}         & 8.9        & 7.0          & {1.9}  & 4.5                                    & 5.2    & 9.4   \\
DiffZO~\cite{magnet2024memory}                & 1.9    & {2.4}   & \textbf{4.2}   & 1.9       & 2.4   & \underline{6.9}  & 2.2         & 3.8        & 2.7        & 2.4   & 4.3              & {4.1}   & -  \\
HybridFMaps~\cite{bastian2024hybrid} & \textbf{1.4}    & 4.2   & 9.5   & 2.3       & \textbf{1.8}   & 13.0 & 2.0        &  4.6       & 3.4      & \textbf{1.8}   & \underline{3.5}                  & \underline{3.9}    & \underline{5.0}  \\

DeepFAFM~\cite{luo2025deep}                & 1.6    & 2.7   & 7.0   & 1.9       & \underline{1.9}   & 7.9 & {2.0}         & 2.9        & \textbf{2.6}        & {1.9}   & {3.9}                    & 4.2    & 6.3 \\
DenoiseFMaps~\cite{zhuravlev2025denoising}           & 1.8    & -   & -   & -      & 2.3   & - & \underline{2.0}         & -       & -        & 2.3   & 4.3      & 12.8  & 43.6 \\

Ours                & \underline{1.6}    & \underline{2.4}  & \underline{4.4}   & \textbf{1.6}        & 2.0     & \textbf{4.0}  & \textbf{1.9}         & \textbf{2.4}      & \textbf{2.4}      & \underline{1.9}   & \textbf{2.9}                                      & \textbf{3.5}   & \textbf{4.9} \\


\bottomrule
\end{tabular}}
\label{tab: near and non-iso}
\end{table}

\textbf{Excluding test-time fine-tuning.} Employing a fine-tuning technique during the inference stage can significantly enhance method performance~\cite{Cao2023}, particularly in terms of generalization capability. However, except for ULRSSM~\cite{Cao2023} and HybridFMaps~\cite{bastian2024hybrid}, the remaining baselines do not adopt any fine-tuning strategies during inference to improve matching performance. Utilizing that refinement technique involving parameter updates during the inference stage would lead to unfair competition with the other unsupervised methods. Therefore, we omit the fine-tuning technique from ULRSSM and HybridFMaps.

\begin{figure*}[h!t]
	\centering
	\includegraphics[width=1.0\linewidth]{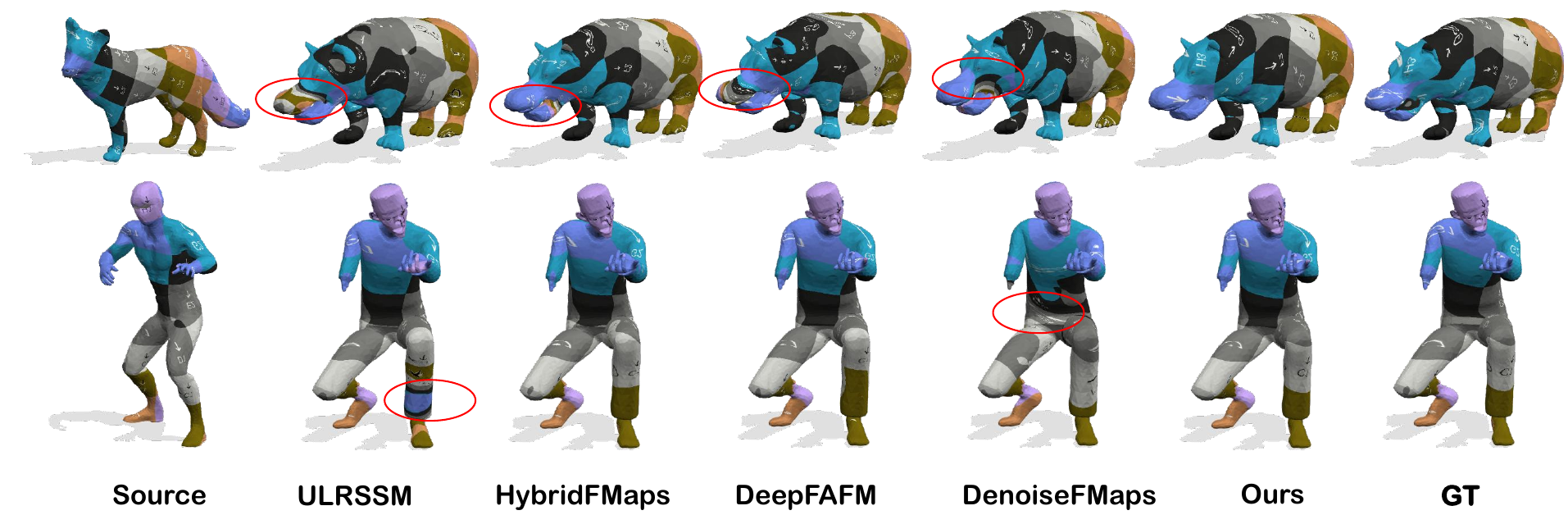}
	\caption{Qualitative non-isometric matching results on the SMAL (top)~\cite{smalzuffi20173d} and DT4D-H (bottom)~\cite{2022SmoothNonRigidShapeMatchingviaEffectiveDirichletEnergyOptimization} datasets. Our method consistently suppresses matching errors and preserves texture coherence better than alternative methods, underscoring its efficacy in challenging non-isometric scenarios.}
	\label{fig:non_iso}
\end{figure*}


\subsection{Results}
The mean geodesic error~\cite{Kim2011BIM} is adopted to evaluate correspondence accuracy, with all results multiplied by 100 for improved readability.

\textbf{Near-isometric matching.} These results are summarized in Table~\ref{tab: near and non-iso}. We achieve competitive performance in near-isometric matching scenarios, surpassing many axiomatic, supervised, and unsupervised methods. For instance, on the FAUST dataset, our method achieves the second-best performance.

\textbf{Cross-dataset generalization.} Cross-dataset generalization is a critical metric for evaluating a method's practical effectiveness. We assess this by training our model on one dataset and testing it on an unseen one, with quantitative results reported in Table~\ref{tab: near and non-iso}. Our method achieves highly competitive performance across most settings. Notably, even on the more challenging SHREC'19 dataset~\cite{melzi2019shrec} (i.e., training on FAUST or SCAPE and testing on SHREC'19), our performance remains robust. In contrast, existing state-of-the-art methods—such as DUO-FMNet~\cite{donati2022deep}, ULRSSM~\cite{Cao2023}, and HybridFMaps~\cite{bastian2024hybrid}—exhibit varying degrees of degradation in relatively simpler cross-dataset scenarios (e.g., training on FAUST, testing on SCAPE) and suffer even steeper performance drops on SHREC'19. These quantitative findings confirm that our approach generalizes substantially better than current baselines.


\textbf{Matching with anisotropic meshing.} To evaluate the robustness of the method under different surface discretizations, we trained on remeshed data and tested on anisotropic remeshed data. The quantitative results are shown in Table~\ref{tab: near and non-iso}. Many state-of-the-art methods exhibited varying degrees of fluctuation. For example, the accuracy of DiffZO significantly dropped when training on FAUST and testing on anisotropic remeshed SCAPE (S\_a). However, our method achieved outstanding performance in most scenarios, demonstrating the robustness of our approach to discretization variations and showing that it consistently outperforms current supervised and unsupervised methods.

\textbf{Non-isometric matching.} We evaluate our method on the highly challenging non-isometric SMAL and DT4D-H datasets, with quantitative results presented in Table~\ref{tab: near and non-iso}. Across all non-isometric settings, our method consistently outperforms all baselines—even fully supervised ones—firmly validating its effectiveness and reliability. In contrast, existing methods such as DenoiseFMaps exhibit a significant drop in accuracy on the particularly demanding DT4D-H dataset. Furthermore, the qualitative results in Fig.~\ref{fig:non_iso} demonstrate that our approach yields significantly lower matching errors and more coherent texture transfers compared to competitive baselines. Together, these results underscore our framework's superior performance and robustness when handling large-scale non-isometric deformations.

\textbf{Matching with topological noise.} We conducted experiments on the SHREC'16 TOPKIDS dataset~\cite{lahner2016shrec} to evaluate robustness against topological perturbations. The results are summarized in Table~\ref{tab: near and non-iso}. Many functional map learning methods~\cite{li2022learning,zhuravlev2025denoising} struggle to produce satisfactory results, as the intrinsic geometric structure is distorted. In contrast, our method still outperforms all baselines and achieves superior performance under such conditions.

For additional comparisons and discussions, please refer to the supplementary materials.




\section{Conclusion}
We propose a novel and effective hybrid self-supervised method for non-rigid shape matching, based on the proposed coarse-to-fine strategy. Extensive experiments demonstrate that our method outperforms existing state-of-the-art methods in terms of matching accuracy, generalization capability, and computational efficiency. Finally, we rigorously demonstrate that contrastive energies promote feature discrimination and yield consistent improvements when integrated into existing architectures, further confirming the validity of our approach.

\section{Acknowledgments}
We extend our gratitude to Yizheng Xie for his generous assistance. 
This work was supported by the National Natural Science Foundation of China under Grant No. 62271452.


%
%
\bibliographystyle{splncs04}
\bibliography{main}

\clearpage
\setcounter{page}{1}

\title{Supplementary Material: Coarse-to-Fine: A Hybrid Self-Supervised Method for Non-rigid 3D Shape Matching} 

\titlerunning{Coarse-to-Fine Hybrid Self-Supervised Learning Shape Matching}

\author{
Feifan Luo\inst{1}\orcidlink{0009-0002-0574-970X} \and
Ting Li\inst{3}\orcidlink{0009-0004-8591-1213} \and
Zhao Li\inst{2}\orcidlink{0000-0002-5056-0351} \and
Hongyang Chen\inst{2}\orcidlink{0000-0002-7626-0162}\thanks{Corresponding author.}
}

\authorrunning{Feifan Luo et al.}

\institute{Zhejiang University, China \and
Zhejiang Lab, China \and Wenzhou University of Technology, China \\
\email{{\tt\small luoff@zju.edu.cn}\quad {\tt\small dr.h.chen@ieee.org}}
}


\maketitle

This supplementary document provides additional theoretical and experimental details to support the findings presented in the main manuscript. First, Section~\ref{sec:the_pro} presents formal proofs for our map refinement and extrinsic energy formulations. Next, Sections~\ref{sec: Imp} through~\ref{sec:runtime} offer a comprehensive experimental evaluation, encompassing implementation details, qualitative results, comparative analyses with fine-tuning-based methods, ablation studies, parameter sensitivity analysis, and runtime comparisons. Subsequently, Section~\ref{sec:method_analysis} provides an in-depth analysis of the proposed method, while Section~\ref{sec:related_to_others} discusses the relationship between our approach and related works. Furthermore, Section~\ref{sec:pro_other_method} demonstrates the versatility of our spectral and spatial-spectral contrastive energies by integrating them into alternative frameworks as plug-and-play modules. Finally, Section~\ref{sec:lim_and_future} addresses current limitations and outlines promising avenues for future research.

\section{Theoretical Proofs}\label{sec:the_pro} 
For the sake of completeness, we provide detailed theoretical derivations in the supplementary material, specifically: (i) the analytical derivation of map refinement, and (ii) the Frobenius norm formulation of the contrastive energies utilized in the extrinsic branch.

\subsection{Map Refinement}
\newtheorem{lem2}[theorem]{Lemma}
\begin{lem2}\label{pro:lemma1}
$E({C}_{\mathcal{XY}}) = \sum^{S}_{s=1} \|   {C}_{\mathcal{XY}}g_{s}(\Lambda_\mathcal{X}) - g_{s}(\Lambda_\mathcal{Y}){C}^{init}_{\mathcal{XY}} \|^{2}_{\mathrm{HS}}$ can be formulated in the Hilbert-Schmidt norm as:
\begin{equation}\label{equ:reg and resolve functional map}
{C}^{ref}_{\mathcal{XY}} = \arg \min_{{C}_{\mathcal{XY}}} \sum^{S}_{s=1} \| {A^{1/2}_{\Y, k}}  \left ({C}_{\mathcal{XY}}g_s(\Lambda_\mathcal{X}) - g_s(\Lambda_\mathcal{Y}){C}^{init}_{\mathcal{XY}} \right ){A^{-1/2}_{\X, k}}\|^{2}_{\mathrm{F}},
\end{equation}
where $A_{\X, k} = \Phi_{\X}^{\mathrm{T}}A_{\X}\Phi_{\X}$.

Moreover, if the set of filter functions $\{g_s(\lambda)\}_{s=1}^S$ satisfy the condition that 
	$\sum_{s} g_s^{2}\left(\lambda\right) \equiv 1, \forall \lambda,$
then the functional map can be obtained via
\begin{equation}\label{eq: solve spectral functional map app}
\begin{aligned}
    {C}^{ref}_\mathcal{XY} = \sum_{s}  g_s\left(\Lambda_{\mathcal{Y}}\right) {C}^{init}_\mathcal{XY}  g_s\left(\Lambda_{\mathcal{X}}\right). 
 \end{aligned}
\end{equation}
\end{lem2}

\begin{proof}
The proof consists of two parts: first, converting the HS-norm in Eq.~\eqref{equ:reg and resolve functional map} to the Frobenius norm, and second, solving for refined functional map ${C}^{ref}_\mathcal{XY}$ from the transformed optimization problem Eq.~\eqref{equ:reg and resolve functional map}.

For the first part, the definition of the Hilbert-Schmidt norm for a general operator on a Hilbert space~\cite{2023ElasticBasis}, namely,
\begin{equation}\label{eq: hs-norm in elastic basis}
\|X\|^{2}_{\mathrm{HS}} =  \mathrm{tr}(X^{*}X), 
\end{equation}
where $\tr$ denotes \textit{trace} and $X^{*}$ represents the adjoint operator of $X$. 

If $X :\mathcal{F(\X)} \to \mathcal{F(\Y)}$, then the following equivalence holds with the Frobenius norm~\cite{2023ElasticBasis,bastian2024hybrid}:
\begin{align}
\label{eq:hs_frob_equiv}
    \|X\|^2_{\mathrm{HS}} :&= \mathrm{tr}(A_{\X,k}^{-1} X^{\T} A_{\Y,k} X ) \\ \nonumber
&= {\tr}({A_{\X,k}^{-1/2}} X^{\T} {A^{1/2}_{\Y,k}}{A^{1/2}_{\Y,k}} X A_{\X,k}^{-1/2}) \\ \nonumber
&= \left\| {A^{1/2}_{\Y,k}} X {A_{\X,k}^{-1/2}} \right\|_\F^2 \nonumber, where \|X\|^2_{\mathrm{F}}=tr(X^{\T}X).
\end{align}
Let $X = {C}_{\mathcal{XY}}g(\Lambda_\mathcal{X}) - g(\Lambda_\mathcal{Y}){C}^{init}_{\mathcal{XY}}$, $s=1,2,...,S$, then we have Eq.~\eqref{equ:reg and resolve functional map}.

For the second part, Eq.\eqref{equ:reg and resolve functional map} can be equivalently transformed into 
$$ 
{A^{1/2}_{\Y, k}} {C}_{\mathcal{XY}}g_s(\Lambda_\mathcal{X}) {A^{-1/2}_{\X, k}} = {A^{1/2}_{\Y, k}} g_s(\Lambda_\mathcal{Y}){C}^{init}_{\mathcal{XY}} {A^{-1/2}_{\X, k}}, s=1,2,...,S.$$
On the other hand, since the inverses of 
${A_{\X, k}}$ and ${A_{\Y, k}}$ exist~\cite{2023ElasticBasis}. Then, by left-multiplying both sides of the equation by 
${A^{-1/2}_{\Y, k}}$
 and right-multiplying by 
${A^{1/2}_{\X, k}}$, the above expression is transformed into:
$$ 
{C}_{\mathcal{XY}}g_s(\Lambda_\mathcal{X}) = g_s(\Lambda_\mathcal{Y}){C}^{init}_{\mathcal{XY}}, s=1,2,...,S.$$

Next, we can obtain the analytical expression of $ {C}^{ref}_{\mathcal{XY}} $ via least square method, the detailed derivation is given below: 
$$\begin{aligned}
	{C}_{\mathcal{XY}}  
	 g_s(\Lambda_\mathcal{X}) & = g_s(\Lambda_\mathcal{Y}){C}_{\mathcal{XY}}, s=1,2,...,S \\
  	{C}_{\mathcal{XY}}  g^{2}_s(\Lambda_\mathcal{X})    &= g_s(\Lambda_\mathcal{Y}){C}_{\mathcal{XY}}g_s(\Lambda_\mathcal{X}),\\
	{C}_{\mathcal{XY}} \sum_{s}  g_s^{2}\left(\Lambda_{\X}\right) & =\sum_{s}   g_s\left(\Lambda_{\Y}\right) {C}_{\mathcal{XY}}  g_s\left( \Lambda_{\X}\right).
\end{aligned}$$
Finally, we have ${C}^{ref}_\mathcal{XY} = \sum_{s}  g_s\left(\Lambda_{\mathcal{Y}}\right) {C}^{init}_\mathcal{XY}  g_s\left(\Lambda_{\mathcal{X}}\right)$, with $\sum_{s} g_s^{2}\left(\lambda\right) \equiv 1, \forall \lambda$.

\end{proof}

\subsection{Extrinsic  Energy}
Since the four contrastive energy expressions in the extrinsic branch have a consistent structure, it is sufficient to prove one of the expressions. 
\newtheorem{lem1app}[theorem]{Lemma}
\begin{lem1app}\label{pro:lemma2}
the bijectivity energy $$ L^{EL}_{bi}  = \left\|\Phi^{EL}_{\mathcal{Y}} - {\Pi}_{\mathcal{YX}}\Phi^{EL}_{\mathcal{X}} \hat{C}^{EL}_\mathcal{YX}\right\|_{\mathrm{HS}},$$ which can be reformulated in terms of the standard Frobenius norm, denoted as:
\begin{equation}\label{equ: formulated elastic bijectivity loss app}
     L^{EL}_{bi}  = \left\|  A^{1/2}_{\Y} \left (  \Phi^{EL}_{\mathcal{Y}} - {\Pi}_{\mathcal{YX}}\Phi^{EL}_{\mathcal{X}} (\hat{C}^{EL}_\mathcal{YX}) \right ) A^{-1/2}_{\Y,k}\right\|_{\mathrm{F}},
\end{equation}
\end{lem1app}

\begin{proof}

If $Y :\mathcal{F(\Y)} \to \mathbb{R}^{|V_{\Y}|}
$, then the following equivalence holds with the Frobenius norm:
\begin{align}
    \|Y\|^2_{\mathrm{HS}} :&= \mathrm{tr}(A_{\Y,k}^{-1} Y^{\T} A_{\Y} Y ) \\ \nonumber
&= {\tr}({A_{\Y,k}^{-1/2}} Y^{\T} {A^{1/2}_{\Y}}{A^{1/2}_{\Y}} Y A_{\Y,k}^{-1/2}) \\ \nonumber
&= \left\| {A^{1/2}_{\Y}} Y {A_{\Y,k}^{-1/2}} \right\|_\F^2 \nonumber.
\end{align}
Let $Y = \Phi^{EL}_{\mathcal{Y}} - {\Pi}_{\mathcal{YX}}\Phi^{EL}_{\mathcal{X}} \hat{C}^{EL}_\mathcal{YX}$, then we obtain Eq.~\eqref{equ: formulated elastic bijectivity loss app}.
\end{proof}

\section{Implementation Details}\label{sec: Imp}
All experiments were executed on a server with Intel(R) Xeon(R) Platinum 8358 CPU @ 2.60GHz, an NVIDIA A40 GPU using PyTorch 2.0.0 and CUDA 11.8. We utilize DiffusionNet~\cite{sharp2022diffusionnet} with its default configurations for feature extraction, employing 16-dimensional Heat Kernel Signatures (HKS)~\cite{sun2009concise} as input to produce a 128-dimensional feature embedding. The spectral resolution is set to $k_{LB}=140$ for the Laplace-Beltrami eigensystem and $k_{EL}=60$ for the Elastic eigensystem. For coarse map computation, we employ a temperature of $\tau = 0.07$, while the refinement stage incorporates Meyer filter functions~\cite{leonardi2013tight} across $S=6$ scales. Notably, to
ensure that the extrinsic loss values are comparable in magnitude to their intrinsic counterparts, the term $A_{{\Y}}^{1/2}$ is omitted from the extrinsic loss formulations. Consequently, the unsupervised loss weights remain predominantly uniform, with all $\theta$ parameters fixed at $1.0$, except for the self-contrastive weights $\theta^{LB/EL}_{self}$, which are set to $0.5$, demonstrating the stability of our framework.


\section{Qualitative Results}\label{sec:qua_results}
\begin{figure*}[h!t]
	\centering
	\includegraphics[width=0.9\linewidth]{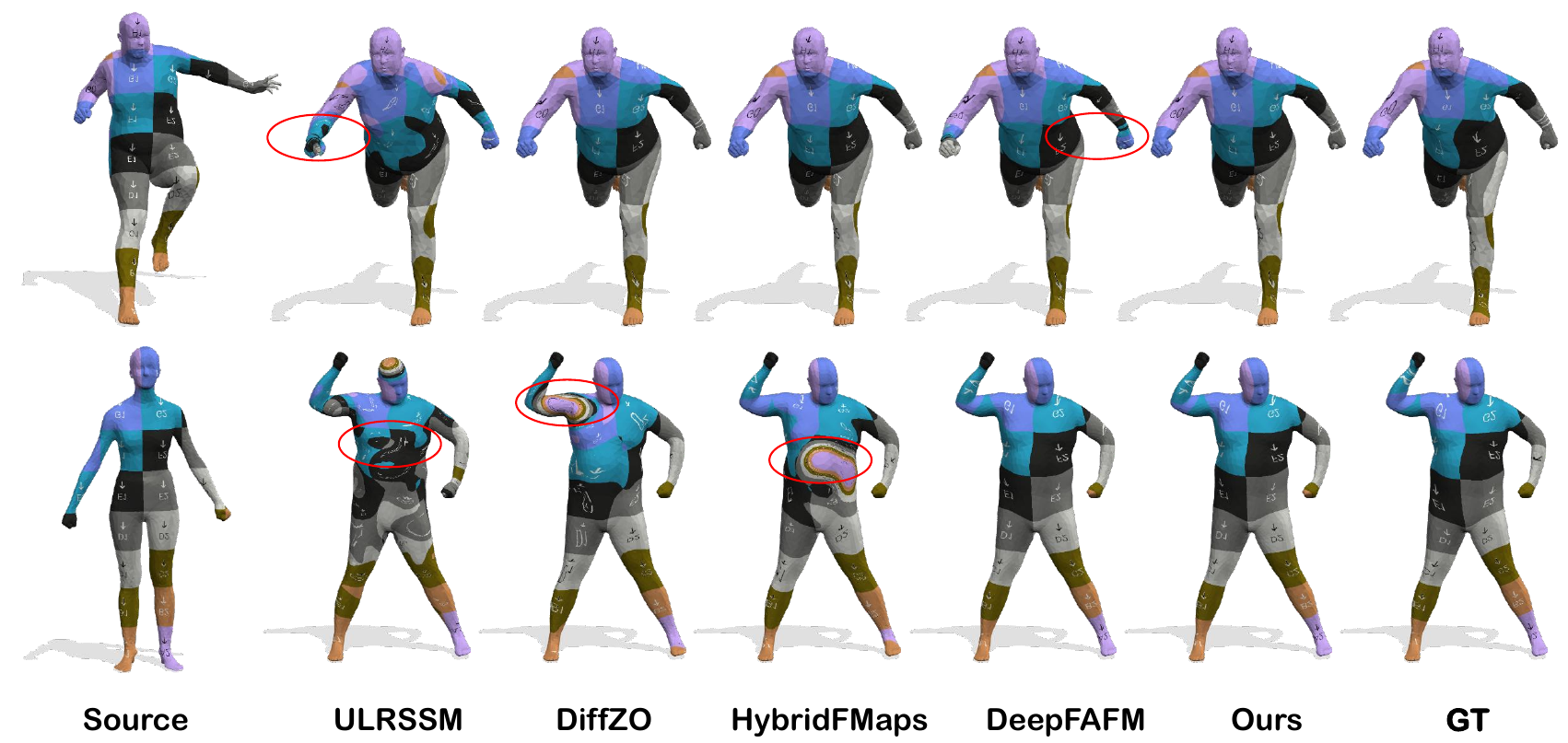}
	\caption{ Qualitative comparison of cross-dataset generalization performance. The top row illustrates results for models trained on FAUST~\cite{ren2018continuous} and evaluated on SHREC’19~\cite{melzi2019shrec}, while the bottom row depicts models trained on SCAPE~\cite{ren2018continuous} and evaluated on SHREC’19. Our method demonstrates superior accuracy with significantly fewer artifacts and more consistent color transfers compared to existing baselines, underscoring its robustness to domain shifts.}
	\label{fig:iso_gen}
\end{figure*}

\begin{figure*}[h!t]
	\centering
	\includegraphics[width=0.95\linewidth]{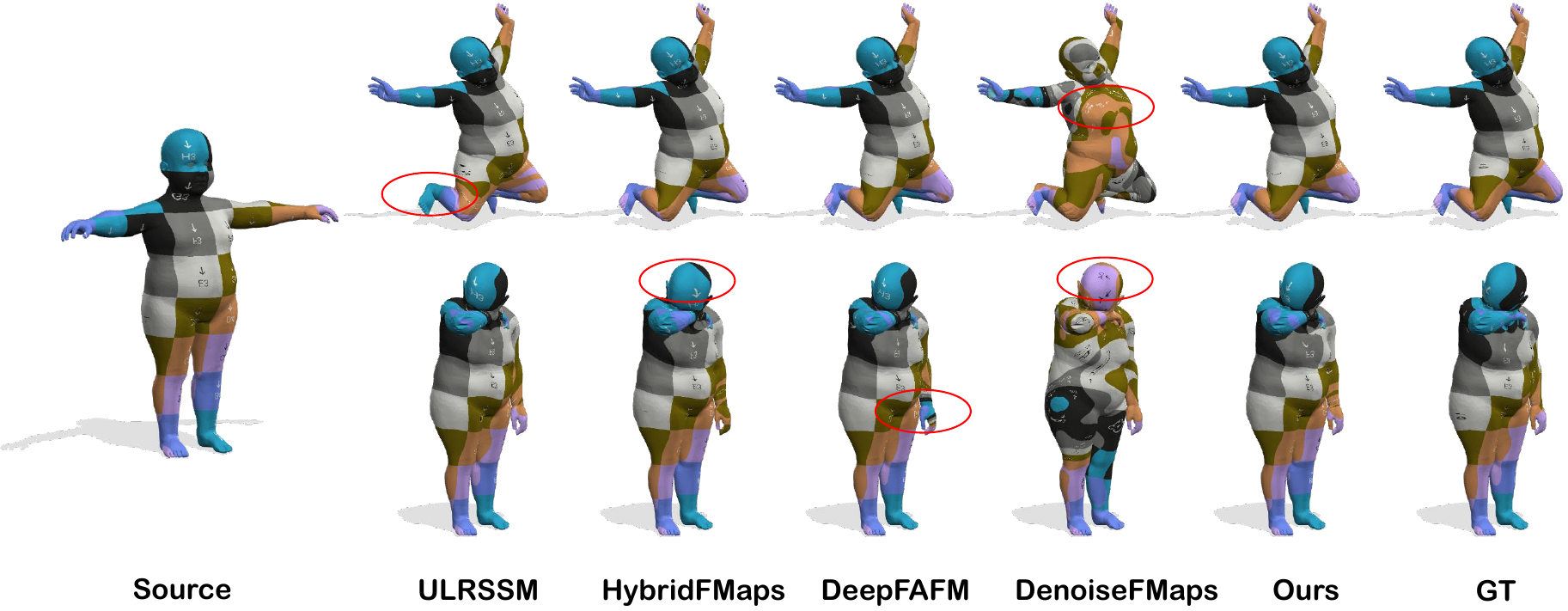}
	\caption{Qualitative comparison of shape matching under topological noise. Results are evaluated on the SHREC’16 Topology benchmark~\cite{lahner2016shrec}. Compared to existing baselines, our method produces more spatially consistent texture transfers with significantly fewer mapping artifacts. This demonstrates the superior robustness of our framework to connectivity inconsistencies and local topological perturbations.
    }\label{fig:topo}
\end{figure*}

In this section, we provide the qualitative results of our method on SHREC'19 (see Fig.~\ref{fig:iso_gen}) and TOPKIDS (see Fig.~\ref{fig:topo}) corresponding to the quantitative results reported in the main text.

\begin{table}[h!t]
\centering
\caption{Comparing our method with fine-tuning approaches~\cite{Cao2023,bastian2024hybrid} across all benchmarks. The numbers in the table are mean geodesic errors ($\times 100$). \textbf{Bold}: Best. \underline{Underline}: Runner-up.}\label{tab:compare_with_finetune}

\scalebox{0.8}{
\begin{tabular}{lccccccccccccc}
    \toprule
Train               & \multicolumn{3}{c}{FAUST} & \multicolumn{3}{c}{SCAPE} & \multicolumn{2}{c}{FAUST} & \multicolumn{2}{c}{SCAPE} & \multicolumn{1}{c}{\multirow{2}{*}{SMAL}}  & \multicolumn{1}{c}{\multirow{2}{*}{DT4D-H inter}} & \multicolumn{1}{c}{\multirow{2}{*}{TOPKIDS}}
\\ \cmidrule(lr){2-4} \cmidrule(lr){5-7} \cmidrule(lr){8-9} \cmidrule(lr){10-11} 
Test                & F        & S   & S19    & F        & S   & S19    & F\_a      & S\_a  & F\_a    & S\_a  & \multicolumn{1}{c}{}                      & \multicolumn{1}{c}{}

\\ \midrule  
                    & \multicolumn{11}{c}{Unsupervised Methods} \\                                                
ULRSSM~\cite{Cao2023}         & {1.6}         & 6.7   & 14.5     &  4.8       & {1.9}   & 18.5   & {2.5}         & 8.9        & 7.0          & {1.9}  & 4.5                                    & 5.2    & 9.4   \\
HybridFMaps~\cite{bastian2024hybrid} & \underline{1.4}    & 4.2   & 9.5   & 2.3       & \underline{1.8}   & 13.0 & 2.0        &  4.6       & 3.4      & \underline{1.8}   & {3.5}                  & {3.9}    & \underline{5.0}  \\
 \midrule
& \multicolumn{11}{c}{Fine-tuning Methods} \\
ULRSSM(w.FT)       & 1.6         & \underline{2.2}   & 5.7     & 1.6         & 1.9   & 6.7  & 1.9         & 2.4        & \underline{2.1}         & 1.9    & 4.2         & 4.1   & 9.2  \\
HybridFMaps(w.FT) & \textbf{1.4}   & \textbf{2.1}   & \underline{5.6}   & \textbf{1.4}       & \textbf{1.8}   & \underline{5.3}  &  \textbf{1.8 }        & \textbf{2.2}        & \textbf{2.1}       & \textbf{1.7}   & \textbf{2.8}           & \underline{3.5}  & 5.1 \\
\midrule
Ours                & {1.6}    & {2.4}  & \textbf{4.4}   & \underline{1.6}        & 2.0     & \textbf{4.0}  & \underline{1.9}         & \underline{2.4}      & {2.4}      & \underline{1.9}   & \underline{2.9}                                      & \textbf{3.5}   & \textbf{4.9} \\

\bottomrule
\end{tabular}}
\end{table}

\section{Compared with the Fine-tuning Methods}\label{sec:com_finetune}
By incorporating those fine-tuning-based methods as baselines, we aim to demonstrate that our approach can achieve outstanding performance without the need for fine-tuning during testing, even surpassing these methods. The comparative results both with and without test-time fine-tuning are summarized in Table~\ref{tab:compare_with_finetune}. 

Our analysis indicates that while fine-tuning yields a comprehensive performance boost for baseline methods, particularly in cross-dataset generalization scenarios, e.g., training on FAUST and testing on SHREC'19. Such gains typically stem from bridging the distributional discrepancies (domain shift) between training and testing sets by iteratively updating parameters to fit the test-time data distribution. In contrast, our framework maintains high precision and robustness to cross-dataset generalization through a strictly feed-forward pass. This suggests that our proposed self-supervised energy terms effectively regularize the model to learn domain-invariant geometric features, eliminating the need for per-pair optimization during inference. Furthermore, in the majority of scenarios involving non-isometric deformations and topological noise, our method outperforms even the fine-tuned results of the baselines. This demonstrates that our framework attains superior accuracy through a purely feed-forward pass during inference, surpassing methods that rely on iterative parameter optimization.

\begin{table}[h!t]
\centering
\caption{We conduct ablation studies on remeshed FAUST, SHREC'19, SMAL, DT4D-H, and TOPKIDS. Where $L^{LB}_{contrast} = L^{LB}_{cross}+L^{LB}_{self}$, $L^{EL}_{contrast} = L^{EL}_{cross}+L^{EL}_{self}$, and $L^{EL}_{fmap} = L^{EL}_{or}+L^{EL}_{bi}$.}
\scalebox{0.95}{
\begin{tabular}{lrrrr}
\hline
\multicolumn{1}{l}{\multirow{2}{*}{Settings}}               & \multicolumn{2}{c}{F} & \multicolumn{1}{c}{\multirow{2}{*}{SMAL}} & \multicolumn{1}{c}{\multirow{2}{*}{TOPKIDS}} \\ \cline{2-3}
            & F       & S19           \\ \hline
w.o $L^{LB}_{contrast}$ and $L^{EL}_{total}$     & 1.7      & 6.2      & 5.3  & 26.8  \\ 
w.o $L^{EL}_{total}$    & 1.6      & \textbf{4.3}      & 3.3  & 12.9  \\
w.o $L^{EL}_{fmap}$      & 1.6    & 5.3    & 3.6  & 5.5  \\ 
w.o $L^{EL}_{contrast}$     & 1.7     & 4.8     & 3.7  & 11.3  \\ 
Ours               & \textbf{1.6}    & 4.4  & \textbf{2.9}   & \textbf{4.9}   \\ 
\hline
\end{tabular}}
\label{tab: ablation studies}
\end{table}

\begin{figure*}[h!t]
	\centering
	\includegraphics[width=0.8\linewidth]{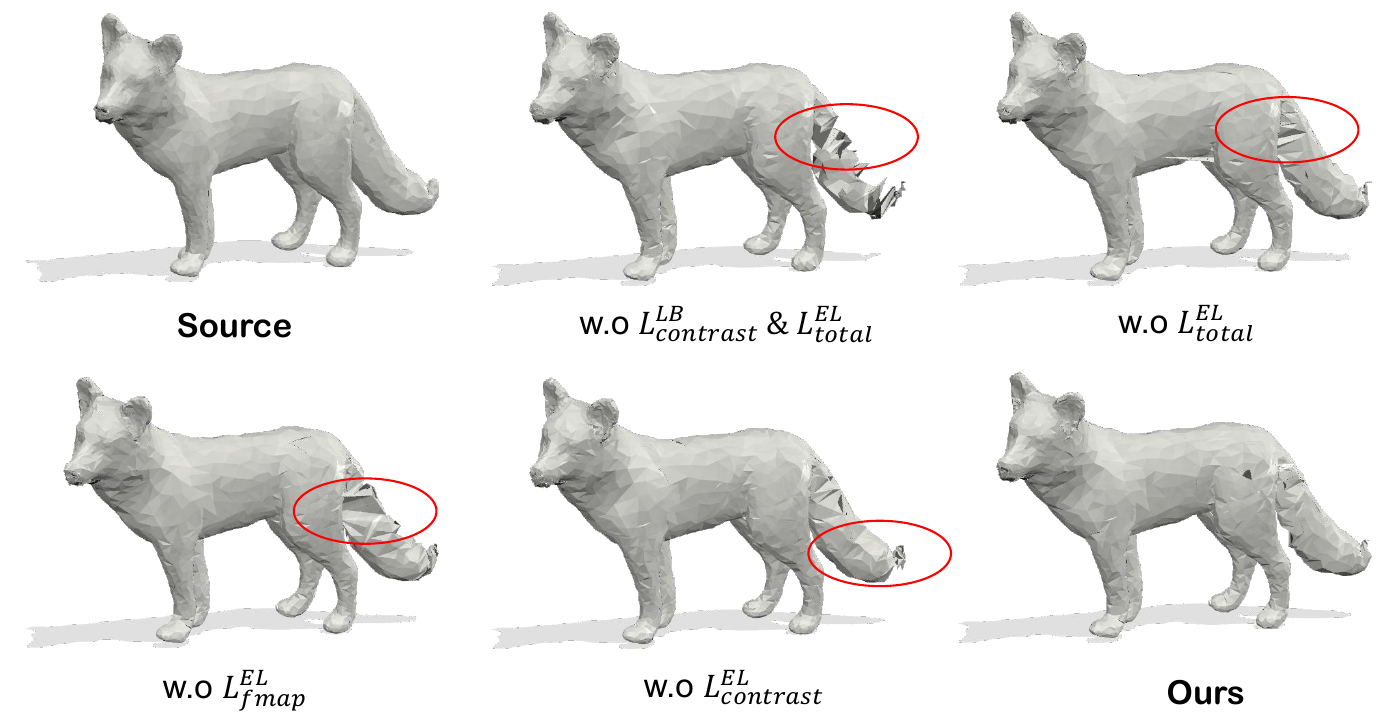}
	\caption{As demonstrated by vertex transfer on the SMAL dataset~\cite{smalzuffi20173d}, the integration of our proposed intrinsic and extrinsic contrastive energies and the extrinsic geometric prior significantly improves matching accuracy and produces visually smoother mappings.}
	\label{fig:alb_std}
\end{figure*}

\section{Ablation Studies}\label{sec:abl_study}
We conduct ablation studies across multiple datasets, covering near-isometric, cross-dataset generalization, non-isometric, and topological noise scenarios, to evaluate the performance of our proposed hybrid self-supervised components: intrinsic contrastive energies $L^{LB}_{contrast}$, extrinsic contrastive energies $L^{EL}_{contrast}$, and extrinsic geometric prior energies $L^{EL}_{fmap}$, respectively.

Our ablation study, summarized in Table~\ref{tab: ablation studies}, highlights the contribution of each component in our framework. On near-isometric benchmarks, we observe that the impact of the hybrid self-supervised components is marginal. This suggests that near-isometric matching is already highly optimized, leaving limited headroom for further improvement. Consequently, our analysis focuses on more challenging scenarios: cross-dataset generalization, non-isometric deformations, and topological noise. 

Comparison between the first and final rows reveals that our proposed energy terms yield comprehensive performance gains, particularly in challenging scenarios involving topological artifacts and non-isometry. A more granular analysis of the second and final rows highlights that the extrinsic energy terms are pivotal for robustness against topological noise; their omission results in a precipitous decline in accuracy on the TOPKIDS dataset. Furthermore, comparing the third and last rows demonstrates that while removing the extrinsic geometric prior energies $L^{EL}_{fmap}$ leads to a marginal decrease in accuracy for non-isometric and topological cases, it causes a substantial degradation in generalization capability. Similarly, the comparison between the fourth and final rows indicates that the absence of extrinsic contrastive energies—specifically $L^{EL}_{contrast}$—results in a minor reduction in generalization but a significant drop in performance under topological noise. Lastly, the comparison between the first and second rows confirms that $L^{LB}_{contrast}$ provides consistent improvements across all evaluated metrics. Collectively, these results underscore the efficacy and necessity of our proposed hybrid self-supervised framework in regularizing complex shape correspondences. 

Moreover, the qualitative results presented in Fig.~\ref{fig:alb_std} demonstrate that our proposed intrinsic and extrinsic contrastive energies, combined with the extrinsic geometric prior, boost matching performance and yield smoother vertex transfers.

\section{Parametric Analysis}\label{sec:para_analysis}

Our framework maintains a minimal hyperparameter configuration. Specifically, the temperature factor ($\tau = 0.07$) and the dimensions of the hybrid eigenbasis ($140$ for the Laplacian basis~\cite{pinkall1993computing} and $60$ for the elastic basis~\cite{2023ElasticBasis}) are kept consistent with recent state-of-the-art work~\cite{bastian2024hybrid}. Consequently, our parametric analysis focuses primarily on the choice of filter functions and the number of scales $S$. These filters can be formulated as orthogonal polynomials~\cite{li2024deformable}, heat kernels~\cite{sun2009concise}, wavelet kernels~\cite{Aubry2011The}, and can even be learned in a data-driven manner~\cite{luo2025deep}. 

\begin{figure*}[h!t]
	\centering
	\includegraphics[width=0.5\linewidth]{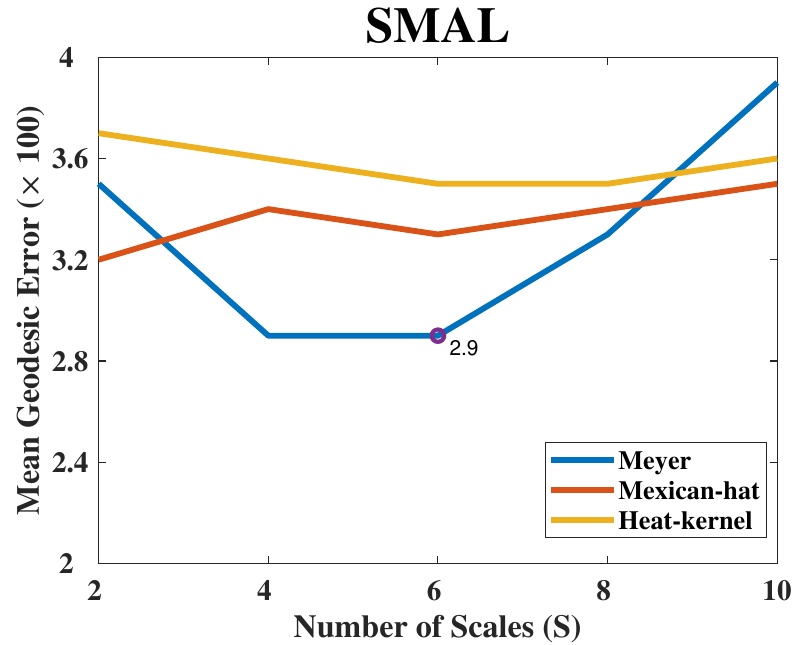}
    \caption{Parametric analysis on the SMAL dataset~\cite{smalzuffi20173d}. We report the mean geodesic error across various filter bank sizes and types, specifically comparing Meyer~\cite{leonardi2013tight}, Mexican-hat~\cite{hammond2011wavelets}, and heat kernel~\cite{sun2009concise} filters (with $t$ uniformly sampled in $[0, 1]$). Empirically, employing 6 Meyer filters emerges as the optimal configuration.}\label{fig:parameter analysis}
\end{figure*}

We report the matching performance of our approach across various types and quantities of filter functions, as illustrated in Fig.~\ref{fig:parameter analysis}. We observe that a set of 6 Meyer filters achieves the optimal configuration; consequently, we adopt this as our default parameter setting.

\section{Runtime Analysis}\label{sec:runtime}
\begin{figure*}[h!t]
	\centering
	  \begin{subfigure}[b]{0.45\textwidth}
    \centering
	\includegraphics[scale=0.35]{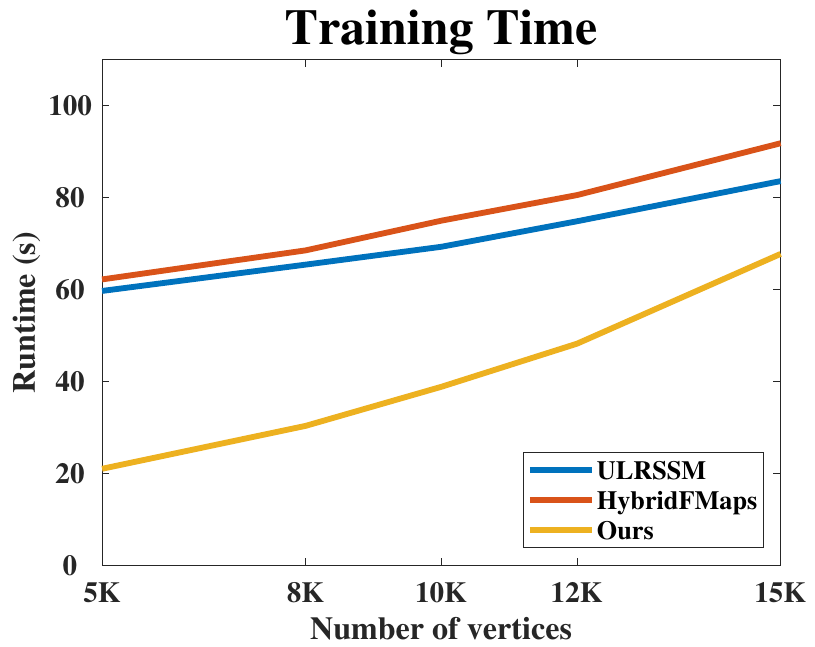}
    \end{subfigure}
    \begin{subfigure}[b]{0.45\textwidth}
    \centering
	\includegraphics[scale=0.35]{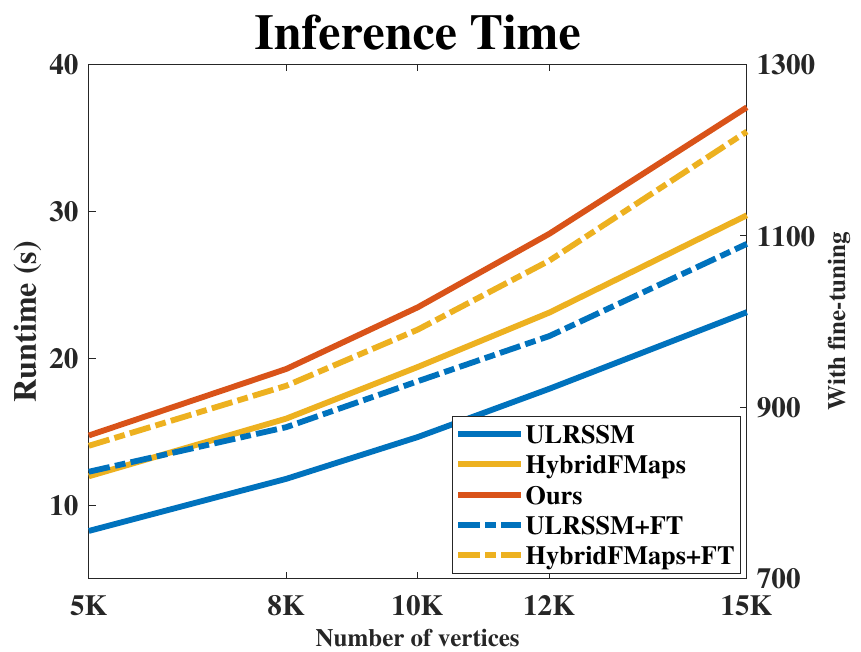}
    \end{subfigure}
    \caption{Runtime analysis across varying vertex resolutions. Left: Mean training duration per method (measured over 100 iterations). Right: Average inference time per 100 shape pairs. Note that the runtimes for ULRSSM(+FT) and HybridFMaps(+FT) (dotted lines) are plotted against the secondary (right) y-axis due to their substantial latency ($\geq$ 50$\times$ higher than other methods); all other baselines correspond to the primary (left) y-axis. Our framework exhibits superior computational efficiency, striking an optimal balance between matching accuracy, training duration, and inference latency.} 
	\label{fig:runtime}
\end{figure*}

The computational efficiency of our method is evaluated across datasets with varying vertex resolutions (5K, 8K, 10K, 12K, and 15K), focusing on both training and inference times. As shown in Fig.~\ref{fig:runtime}(left), our training speed significantly outperforms state-of-the-art methods, including ULRSSM~\cite{Cao2023} and HybridFMaps~\cite{bastian2024hybrid}. Specifically, ULRSSM solves $k$ independent $k \times k$ linear systems via least squares, yielding a time complexity of $O(k^4)$, while HybridFMap imposes a substantially higher computational burden. Beyond the spectral functional map computation on the Laplacian basis, HybridFMap necessitates solving a $(k^{EL})^2 \times (k^{EL})^2$ system derived from the elastic basis. This results in an overall complexity of $O((k^{LB})^4 + (k^{EL})^6)$. Since the $O((k^{EL})^6)$ term constitutes a severe bottleneck, HybridFMap is forced to utilize a strictly limited number of elastic basis functions ($k^{EL}=60$) to avoid prohibitive overhead, thereby maintaining a total execution time comparable to that of ULRSSM.

In contrast, our approach substantially reduces computational costs by eliminating the least-squares solver typically used for the spectral functional map computation on both Laplacian and elastic bases. Furthermore, while existing methods require computing bidirectional spectral and spatial functional maps to accommodate their bidirectional loss formulations, our framework only necessitates the unidirectional spatial functional maps, ensuring superior efficiency.

Quantitative statistics for inference time are provided in Fig.~\ref{fig:runtime} (right). Our method achieves an inference speed comparable to that of ULRSSM and HybridFMaps, primarily because our additional refined map calculation introduces only negligible overhead. It is worth noting that while ULRSSM and HybridFMaps report competitive speeds, they heavily rely on test-time adaptation, a fine-tuning technique that iteratively updates network parameters during inference (e.g., 15 backpropagation iterations per shape pair). For a standard set of 100 pairs, this results in an additional 1500 iterations for parameter optimizations, significantly increasing the computational burden. In contrast, our method achieves superior performance without any test-time parameter updating, even outperforming baseline methods that utilize such a fine-tuning technique, as shown in Table~\ref{tab:compare_with_finetune}.


\section{Method Discussion}\label{sec:method_analysis}
Our unsupervised loss functions are comprised of two distinct components: axiomatic prior-based energies and contrastive energies. While the former enforces fundamental geometric constraints for functional maps, such as orthogonality (or local area preservation~\cite{rustamov2013map}) and bijectivity. The latter is specifically designed to enhance feature representation. In particular, our spatial contrastive terms promote \textit{orthogonality} within the feature space, distinguishing them from traditional geometric regularizers. 

\newtheorem{thm1}{Remark}
\begin{thm1}
In the feature space, these spatial contrastive energy terms act as surrogates for orthogonality constraints within the feature space, namely, 
\begin{equation}
   \| \Pi_{\YX} - \Pi^{ref}_{\YX} \|  \approx \| F_{\Y}F^{\T}_{\X} - I_{\YX} \|
\end{equation}
and
\begin{equation}
  \| \Pi_{\Y\Y} - I_{\Y} \|  \approx \| F_{\Y}F^{\T}_{\Y} - I_{\Y}\|,
\end{equation}
where $\Pi_{\YX} = \mathrm{Softmax}(F_\Y F_\X^\top / \tau)$, $I_{\YX} \in \mathbb{R}^{|V_\Y| \times|V_\X|}$ with each row being a one-hot vector. $I_\Y$ is defined as the identity matrix of size $|V_\Y|$.
\end{thm1}

\begin{proof}
The cross-spatial contrastive energy aims to align the coarse soft assignment map $\Pi_{\YX}$ with the refined hard map $\Pi^{ref}_{\YX}$. Specifically, the soft map is defined by the similarity of vertex features as $\Pi_{\YX} = \mathrm{Softmax}(F_\Y F_\X^\top / \tau)$. Given that the softmax operator normalizes the feature product into a probability distribution, we conceptually approximate the soft map as $\Pi_{\YX} \approx F_\Y F_\X^\top$ to simplify our derivation. Furthermore, since $\Pi^{ref}_{\YX}$ is an indicator matrix, encouraging $F_\Y F_\X^\top$ to approximate $\Pi^{ref}_{\YX}$ effectively enforces cross-orthogonality between the feature sets $F_\X$ and $F_\Y$. In this ideal configuration, each vertex feature $[F_\Y]_{j} \in \mathbb{R}^{d}$ ($j$-th row of $F_\Y$, $j=1,2,...,|V_\Y|$) corresponds to a unique index $i_0$ such that their inner product satisfies the Kronecker delta property: $[F_\Y]_{j} [F_\X]_{i}^\top = \delta_{i, i_0}$, i.e.,  $[F_\Y]_{j} [F_\X]_{i}^\top = 1$ if $i = i_0$, and $[F_\Y]_{j} [F_\X]_{i}^\top = 0$ otherwise. Analogously, the self-contrastive term enforces self-orthogonality among the features, thereby suppressing feature collapse and preserving the uniqueness of individual vertex representations.
\end{proof}

In summary, the cross- and self-spatial contrastive energies provide complementary constraints that collectively enhance the discriminative power of the learned features from both inter-shape and intra-shape perspectives.

\begin{table}[h!t]\label{sec:relation_to_others}
\centering
\caption{Summary of contrastive energies.}
\scalebox{0.90}{
\begin{tabular}{lcl}
\toprule
Formula     &         & Description           \\ \midrule
$\| \Pi_{\YX} - \Pi^{ref}_{\YX} \|$  &   & cross-spatial contrastive energy      \\ 
$\| \Pi_{\Y\Y} - I_{\Y} \|$  &  & self-spatial contrastive energy    \\
$\| C_{\X\Y} - C^{ref}_{\X\Y} \|$ &  & spectral contrastive energy  \\
$\left\| {C}_\mathcal{XY} - \Phi_\mathcal{Y}^\dagger\Pi^{ref}_\mathcal{YX}\Phi_\mathcal{X}\right\|$ &  & spatial-spectral contrastive or fine properness energy        \\
$\left\| \Pi_\mathcal{YX} - \Phi_\mathcal{Y}{C}^{ref}_\mathcal{XY}\Phi^\dagger_\mathcal{X} \right\| $ &  & spatial-spectral contrastive energy induced by refined functional maps \\
\bottomrule
\end{tabular}}
\label{tab: energy desc}
\end{table}

\section{Relation to Other Techniques}\label{sec:related_to_others}
In this section, we demonstrate that the consistency constraint proposed in DiffZO~\cite{magnet2024memory} and the vertex-wise contrastive energy introduced in RevisitedFM~\cite{cao2024revisiting} are, in fact, special cases of our generalized contrastive framework. 

Specifically, DiffZO employs a consistency loss $\|\hat{C}_{\XY} - \hat{C}^{ref}_{\XY}\|$, where the refined spectral map $\hat{C}^{ref}_{\XY}$ is obtained via ZoomOut refinement~\cite{Melzi2019}. This objective can be interpreted as a specific instance of spectral projection of our cross-spatial contrastive energy, formulated as $\|\Phi_{{\Y}}^\dagger \Pi_{\YX} \Phi_{\mathcal{\X}} - \Phi_{{\Y}}^\dagger \Pi^{ref}_{\YX} \Phi_{{\X}}\|$. Similarly, the vertex-wise contrastive energy $\|\hat{C}_{\Y\Y} - I\|$ in RevisitedFM~\cite{cao2024revisiting} represents a spectral-domain specialization of our self-spatial contrastive loss, which can be expressed as $\|\Phi_{{\Y}}^\dagger \Pi_{\Y\Y} \Phi_{{\Y}} - \Phi_{{\Y}}^\dagger \Pi^{ref}_{\Y\Y} \Phi_{{\Y}}\|$. By unifying these diverse objectives, our framework provides a more comprehensive perspective on spatial coarse-to-fine contrast, opening new avenues for future research.

\section{Contrastive Energies in Different Frameworks}\label{sec:pro_other_method}

To prioritize computational efficiency, we intentionally bypassed the use of least-squares solvers for functional map computation in our main pipeline. However, this design choice means that our proposed contrastive energies, specifically, the spectral and spatial-spectral contrastive terms, i.e, $    \left\| {C}_\mathcal{XY} - {C}^{ref}_\mathcal{XY}\right\|$ and     $\left\| {C}_\mathcal{XY} - \Phi_\mathcal{Y}^\dagger\Pi^{ref}_\mathcal{YX}\Phi_\mathcal{X}\right\|$, are not utilized to their full potential within our own framework. To rigorously validate the efficacy of these energy terms and demonstrate their consistent performance gains, we integrated them as plug-and-play modules into two established frameworks: SURFMNet~\cite{roufosse2019unsupervised} and ULRSSM~\cite{Cao2023}.

Specifically, we augmented SURFMNet (which relies on orthogonality and bijectivity losses) with our proposed spectral contrastive loss (corresponding weight $\theta=5$), denoting this variant as SURFMNet*. Similarly, for ULRSSM, we replaced the standard coarse properness loss $\left\| {C}_\mathcal{XY} - \Phi_\mathcal{Y}^\dagger\Pi_\mathcal{YX}\Phi_\mathcal{X}\right\|$ with our fine properness loss, resulting in ULRSSM*. Both the refined pointwise correspondences and functional maps used during the training and inference stages are derived from our proposed map refinement algorithm. Moreover, all remaining hyperparameter configurations and loss weighting coefficients were kept consistent with the original implementations to ensure a rigorous and fair comparison. The quantitative results of this integration are summarized in Table~\ref{tab: coarse-to-fine promote other frameworks}.

Experimental results show that performance consistently improves after adding these regularization terms. This is particularly evident in cross-dataset generalization and topological noise scenarios. In the case of ULRSSM* on SMAL, we observe a slight performance decline compared to ULRSSM, which could be due to the hyperparameters not being tailored to the specific characteristics of the method. Nevertheless, our contrastive energy still brings a substantial improvement to the results of ULRSSM.

In conclusion, incorporating our self-supervised energies provides a comprehensive improvement to these frameworks, highlighting their effectiveness in enhancing both functional and pointwise map representations.

\begin{table}[h!t]
\centering
\caption{Evaluating the matching results across various conditions. The numbers in the table are mean geodesic errors ($\times 100$). \textit{Improvement} refers to the performance enhancement($\uparrow$) achieved after adding the contrastive energy constraints.}
\scalebox{0.9}{
\begin{tabular}{lccccc}
\toprule
\multicolumn{1}{l}{\multirow{2}{*}{Settings}}               & \multicolumn{2}{c}{FAUST} & \multicolumn{1}{c}{\multirow{2}{*}{SMAL}} & \multicolumn{1}{c}{\multirow{2}{*}{DT4D-H}} & \multicolumn{1}{c}{\multirow{2}{*}{TOPKIDS}} \\ \cline{2-3}
            & FAUST       & SHREC'19           \\ \hline
SURFMNet    & 2.5        & 17.6     & 21.3  & 26.1 & 40.7  \\ 
SURFMNet*   & 1.9       & 6.5      & 16.2  & 12.7  & 21.1 \\
\textit{Improvement}($\uparrow$)    & 24\%       & 63\%      & 31\%  & 51\%  & 48\% \\ \hline 
ULRSSM      & 1.6    & 14.5  & 4.5 & 5.2 & 9.4 \\ 
ULRSSM*     & 1.6     & 7.1   & 4.6  & 4.5   & 6.5 \\ 
\textit{Improvement}($\uparrow$)   & 0\%       & 51\%     & -2\%  & 13\%  & 31\% \\
\bottomrule
\end{tabular}}
\label{tab: coarse-to-fine promote other frameworks}
\end{table}

\section{Limitation and Future Work}\label{sec:lim_and_future}
Despite its robustness to challenging scenarios such as non-isometric deformations and topological noise, our method currently struggles with partial shape matching. This limitation arises because the missing boundary regions in partial geometries introduce severe intrinsic distortions, which lead to suboptimal matching performance.

Several promising directions remain for future exploration, such as (1) Enhanced Refinement: Integrating spatial deformation constraints~\cite{eisenberger2020smooth,eisenberger2020deep,eisenberger2021neuromorph,cao2024spectral} or learning filter functions~\cite{luo2025deep} into the refinement stage to better accommodate scenarios characterized by large-scale non-isometric deformations and severe topological noise.
(2) Contrastive Frameworks: Developing a novel matching architecture that leverages spatial-spectral contrastive energy to steer coarse pointwise optimization via refined functional maps, see Table~\ref{tab: energy desc};
(3) Generative Paradigms: Exploring diffusion models~\cite{pierson2025diffumatch,zhuravlev2025denoising} and flow matching~\cite{lipman2022flow,chen2024flow,olearo2025fuse} to bolster generalization across diverse and out-of-distribution datasets;
(4) Robust Shape Operators: Constructing hybrid geometric operators by aggregating multiple spectral bases~\cite{choukroun2018hamiltonian,wang2019intrinsic,weber2024finsler} to improve map stability under noise and topological changes.


%
%

\end{document}